\documentclass[acmtog, authorversion]{acmart}

\makeatother

\acmSubmissionID{915}
\usepackage{graphicx}
\usepackage{booktabs}

\usepackage{multirow}
\usepackage{makecell}
\usepackage{arydshln}
\usepackage{enumitem}
\AtBeginDocument{%
  }

\setcopyright{acmlicensed}
\acmJournal{TOG}
\copyrightyear{2024}
\acmYear{2024} \acmVolume{43} \acmNumber{6} \acmArticle{} \acmMonth{12}
\acmDOI{10.1145/3687975}

\begin{document}
% \begin{sloppypar}

%\title{StyleCrafter: Enhancing Stylized Text-to-Video Generation with Style Adapter}
\title{StyleCrafter: Taming Stylized Video Diffusion with Reference-Augmented Adapter Learning}

\author{Gongye Liu}
\affiliation{
  \institution{Tsinghua University}
  \city{Shenzhen}
  \country{China}
}
\email{lgy22@mails.tsinghua.edu.cn}

\author{Menghan Xia}
\authornote{Corresponding authors}
\author{Yong Zhang}
\author{Haoxin Chen}
\affiliation{
  \institution{Tencent AI Lab}
  \city{Shenzhen}
  \country{China}
}
\email{menghanxyz@gmail.com}
\email{zhangyong201303@gmail.com}
\email{jszxchx@gmail.com}

\author{Jinbo Xing}
\affiliation{%
  \institution{The Chinese University of Hong Kong}
  \city{Hong Kong}
  \country{China}
}
\email{jbxing@cse.cuhk.edu.hk}

\author{Yibo Wang}
\affiliation{%
  \institution{Tsinghua University}
  \city{Shenzhen}
  \country{China}
}
\email{wyb22@mails.tsinghua.edu.cn}

\author{Xintao Wang}
\author{Ying Shan}
\affiliation{%
  \institution{Tencent AI Lab}
  \city{Shenzhen}
  \country{China}
}
\email{xintao.alpha@gmail.com}
\email{yingsshan@tencent.com}

\author{Yujiu Yang}
\authornotemark[1]
\affiliation{%
  \institution{Tsinghua University}
  \city{Shenzhen}
  \country{China}
}
\email{yang.yujiu@sz.tsinghua.edu.cn}


\begin{abstract}

Text-to-video (T2V) models have shown remarkable capabilities in generating diverse videos. However, they struggle to produce user-desired artistic videos due to (i) text's inherent clumsiness in expressing specific styles and (ii) the generally degraded style fidelity. To address these challenges, we introduce StyleCrafter, a generic method that enhances pre-trained T2V models with a style control adapter, allowing video generation in any style by feeding a reference image.
Considering the scarcity of artistic video data, we propose to first train a style control adapter using style-rich image datasets, then transfer the learned stylization ability to video generation through a tailor-made finetuning paradigm. 
To promote content-style disentanglement, we employ carefully designed data augmentation strategies to enhance decoupled learning. Additionally, we propose a scale-adaptive fusion module to balance the influences of text-based content features and image-based style features, which helps generalization across various text and style combinations.
StyleCrafter efficiently generates high-quality stylized videos that align with the content of the texts and resemble the style of the reference images. Experiments demonstrate that our approach is more flexible and efficient than existing competitors.
Project page: \url{https://gongyeliu.github.io/StyleCrafter.github.io/}

\end{abstract}

\begin{CCSXML}
<ccs2012>
 <concept>
  <concept_id>00000000.0000000.0000000</concept_id>
  <concept_desc>Do Not Use This Code, Generate the Correct Terms for Your Paper</concept_desc>
  <concept_significance>500</concept_significance>
 </concept>
 <concept>
  <concept_id>00000000.00000000.00000000</concept_id>
  <concept_desc>Do Not Use This Code, Generate the Correct Terms for Your Paper</concept_desc>
  <concept_significance>300</concept_significance>
 </concept>
 <concept>
  <concept_id>00000000.00000000.00000000</concept_id>
  <concept_desc>Do Not Use This Code, Generate the Correct Terms for Your Paper</concept_desc>
  <concept_significance>100</concept_significance>
 </concept>
 <concept>
  <concept_id>00000000.00000000.00000000</concept_id>
  <concept_desc>Do Not Use This Code, Generate the Correct Terms for Your Paper</concept_desc>
  <concept_significance>100</concept_significance>
 </concept>
</ccs2012>
\end{CCSXML}

\ccsdesc[500]{Computing methodologies~Computer Vision}

\keywords{Diffusion Model, Stylized Generation, Image/Video Synthesis}

\begin{teaserfigure}
  \includegraphics[width=\textwidth]{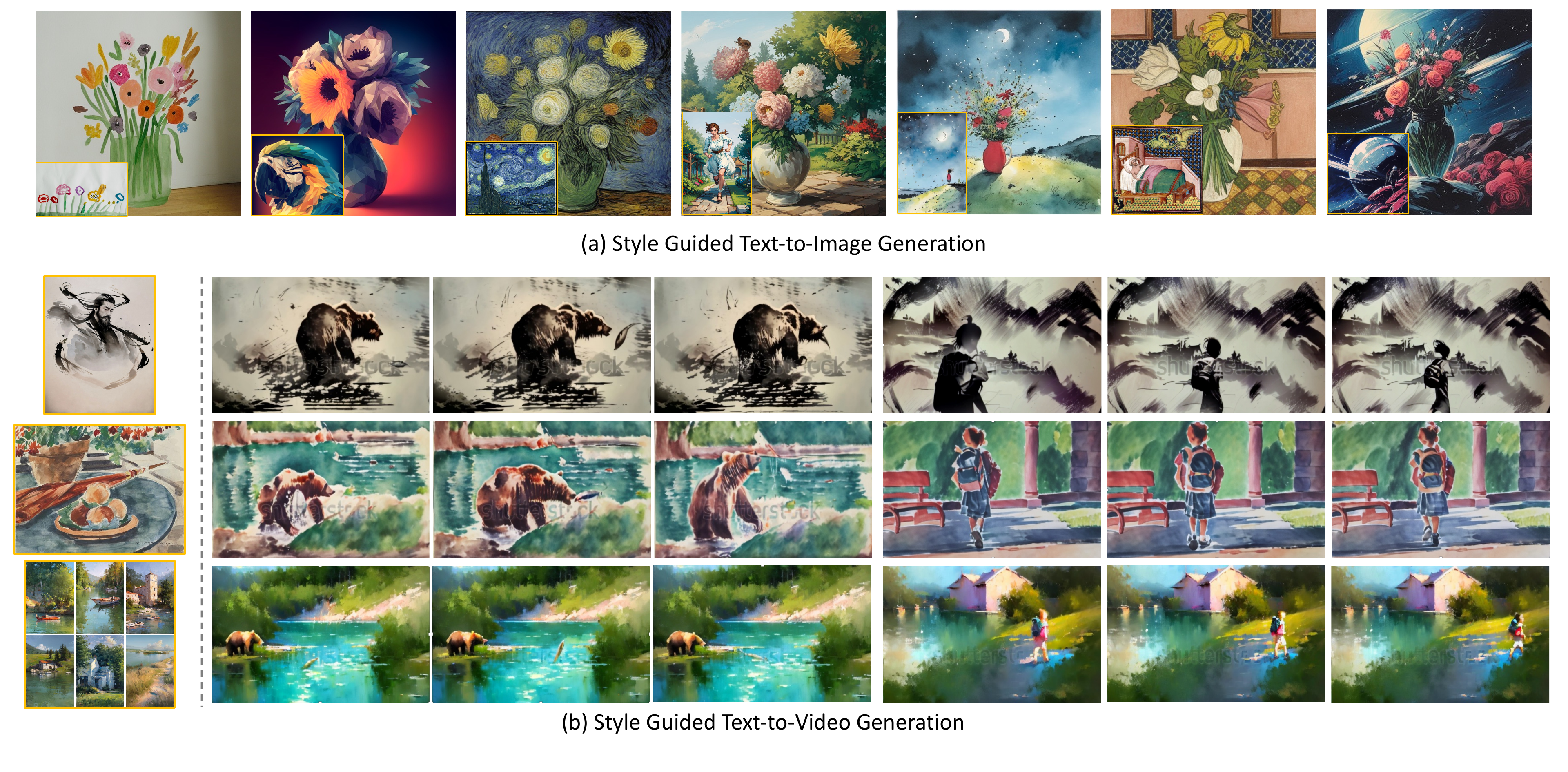}
  \vspace{-3em}
  \caption{Stylized Generation Results Produced by StyleCrafter}
  \Description{Stylized Generation Results Produced by StyleCrafter}
  % \vspace{-2em}
  \label{fig:teaser}
\end{teaserfigure}


\maketitle

\begin{sloppypar}

%%%%%% for TOG %%%%%% 
% % \begin{bibunit}
% \input{sections/01_introduction}
% \input{sections/02_relatedworks}
% \input{sections/03_method}
% \input{sections/04_experiment}
% \input{sections/05_conclusion}

% \citestyle{acmauthoryear}
% \bibliographystyle{ACM-Reference-Format}
% \bibliography{reference}
% \end{bibunit}

% \begin{bibunit}
% \clearpage
% \input{supp}

% \clearpage
% \citestyle{acmauthoryear}
% \bibliographystyle{ACM-Reference-Format}
% \bibliography{reference}
% \end{bibunit}
%%%%%% for TOG %%%%%% 

%%%%%% for arxiv %%%%%% 
%---------------------------------
\section{Introduction}
\label{sec:intro}
%---------------------------------

The popularity of powerful diffusion models has led to remarkable progress in the field of content generation. For instance, text-to-image (T2I) models are capable of generating diverse and vivid images from text prompts, encompassing various visual concepts. This great success can be attributed not only to the advancement of models but also to the availability of various image data over the Internet.
Constrastingly, text-to-video (T2V) models fall short of the data categories especially in styles, since existing videos predominantly feature photorealism. While these strategies, like initializing weights from well-trained T2I models or joint training with image and video datasets, can help mitigate this issue, the generated stylized videos generally suffer from degraded style fidelity. 
% Therefore, in addition to the classic problem of \textbf{style-content decoupling} in style transfer/preserving, stylized video generation also grapples with challenges including \textbf{a scarcity of stylized video data} and the \textbf{limited capabilities of T2V base models}.
Although significant success has been achieved in style transfer/preservation in T2I generation, the field of stylized video generation remains largely unexplored,
and effective solutions are yet to be discovered.
% As one of the pioneers, AnimateDiff~\cite{guo2023animatediff} can make impressive stylized videos by combining personalized T2I models~\cite{hu2022lora} (i.e. LoRA-tuned\cite{hu2022lora} or Dreambooth-tuned\cite{dreambooth} on Stable Diffusion\cite{ldm}) with pre-trained temporal blocks. However, each style requires additional finetuning on a small set of examples, which is inefficient and unable to support any style.

% \begin{figure}[!t]
%     \centering
%     \includegraphics[width=\linewidth]{figures/motivation.pdf}\vspace{-0.5em}
%     \caption{Effect of adding style adapter to T2V models. (a) and (b) are results of Stable Diffusion~\cite{ldm} and VideoCrafter~\cite{chen2023videocrafter}. (c) is the result of VideoCrafter equipped with a style adapter. The content text prompt is "\textit{A knight riding a horse through the field}". For (a) and (b), the style prompt is generated from the style image using GPT4V~\cite{openai2023gpt4v}. \TODO{delete this figure}} 
%     \label{fig:motivation}\vspace{-0.5em}
% \end{figure}

In this paper, we propose StyleCrafter, a generic method that enhances pre-trained T2V models with a style control adapter, enabling text-to-video generation in any desired style by providing a reference image. 
% The advantages are twofold: (i) a style image offers stylistic feature guidance, complementing the stylization capabilities of T2V models in a zero-shot fashion; (ii) the reference image delivers a more accurate portrayal of the desired visual style compared to textual descriptions. 
Anyhow, it is non-trivial to achieve this goal. (i) as a classic problem of style transfer/preservation, the style control adapter requires to extract accurate style concepts from the reference image \textbf{in a content-style decoupled manner}. (ii) \textbf{the scarcity of open-source stylized videos} challenges the adaptation training of the T2V models.

Considering the scarcity of stylized videos, we propose to first train a style adapter to extract desired style concepts from images over image datasets, and then transfer the learned stylization ability to a T2V model with shared spatial weights through a tailor-made finetuning paradigm. The advantages are twofold: on the one hand, the adapter trained over stylized images can effectively extract the style concept from input images, eliminating the necessity for scarcely available stylized videos. On the other, a finetuning paradigm enables text-to-video models with better adaptation to the style concepts extracted from the previously trained style adapter, while avoiding degradation of temporal quality in video generation. 

To effectively capture the style features and promote content-style disentanglement, we adopt the widely used query transformer to extract style concepts from a single image. Particularly, we design a scale-adaptive fusion module to balance the influences of text-based content features and image-based style features, which helps generalization across various text and style combinations. During the training process, we employ carefully designed data augmentation strategies to enhance decoupled learning.

StyleCrafter efficiently generates high-quality stylized videos that align with the content of the texts and resemble the style of the reference images.
Comprehensive experiments are conducted to assess our proposed approach, demonstrating that it significantly outperforms existing competitors in both stylized image generation and stylized video generation. Furthermore, ablation studies offer a thorough analysis of the technical decisions made in developing the complete method, which provides valuable insights for the community.
Our contributions are summarized as follows:
\begin{itemize}
    \item We propose the concept of improving stylized generation for pre-trained T2V models by adding a style adapter.
    \item We explore an efficient network for stylized generation, which facilitates the content-style disentangled generation from text and image inputs. Our method attains notable advantages over existing baselines.
    \item We propose a training paradigm for generic T2V style adapter without requiring any stylized videos for supervision.
\end{itemize}
%---------------------------------
% \vspace{-0.5em}
\section{Related Works}
\label{sec:realtedworks}
%---------------------------------

\subsection{Text to Video Synthesis}
Text-to-video synthesis~(T2V) is a highly challenging task with significant application value, aiming to generate corresponding videos from text descriptions. Various approaches have been proposed, including autoregressive transformer~\cite{vaswani2017attention} models and diffusion models~\cite{DDPM, DDIM, nichol2021improved, song2020score}. 
% N{\"u}wa~\cite{wu2022nuwa} introduces a 3D transformer encoder-decoder framework to address various text-guided visual tasks including T2V generation.  Phenaki~\cite{villegas2022phenaki} presents a bidirectional masked transformer for compressing videos into discrete tokens, thereby enabling video generation. 
Video Diffusion Model~\cite{ho2022video} employs a space-time factorized U-Net to execute the diffusion process in pixel space. Imagen Video~\cite{ho2022imagen} proposes a cascade diffusion model and v-parameterization to enhance VDM. 
Another branch of techniques makes good use of pre-trained T2I models and further introduces some temporal blocks for video generation extension. CogVideo~\cite{hong2022cogvideo} builds upon CogView2~\cite{ding2022cogview2} and employs multi-frame-rate hierarchical training strategy to transition from T2I to T2V. Similarly,  Make-a-video~\cite{make-a-video}, MagicVideo~\cite{zhou2022magicvideo} and LVDM~\cite{he2022latent} inherit pretrained T2I diffusion models and extend them to T2V generation by incorporating temporal attention modules.
%Furthermore, some wrok have explored introducing additional control conditions in T2V diffusion models. Gen-1~\cite{esser2023structure} proposes a structure and content-guided VDM that utilizes frame-wise depth map to maintain structure. VideoComposer~\cite{wang2023videocomposer} focuses on video generation conditioned on multi-modal inputs, allowing textual, spatial, and temporal conditions.
%Follow Your Pose~\cite{ma2023follow} aims to generate pose-controllable character videos by employing a two-stage training process that exclusively utilizes image-pose and pose-free video. Nevertheless, example-based stylized video generation is seldom explored in the general video synthesis field.

\vspace{-0.7em}
\subsection{Stylized Image Generation}

Stylized image generation aims to create images that exhibit a specific style. Decoupling style and content is a classic challenge~\cite{tenenbaum2000separating}.
Early research primarily concentrated on image style transfer, a technique that involves the transfer of one image's style onto the content of another, requiring a source image to provide content. 
Traditional style transfer methods~\cite{hertzmann2001image,wang2004efficient, zhang2013style} employ low-level, hand-crafted features to align patches between content images and style images. Since Gatys et al.~\cite{gatys2016image} discovered that the feature maps in CNNs capture style patterns effectively, a number of studies~\cite{huang2017arbitrary, li2017universal, texler2020arbitrary, liu2021adaattn, an2021artflow, deng2022stytr2, zhang2022domain} have been denoted to utilize neural networks to achieve arbitrary style transfer. A common practice involves utilizing a pretrained VGG network~\cite{simonyan2014very} to extract style information or compute Gram matrix loss~\cite{gatys2016image} to enable self-supervised learning of visual styles.

As the field of generation models progressed, researchers began exploring stylized image generation for T2I models. Although T2I models can generate various artistic images from corresponding text prompts, words are often limited to accurately convey the stylistic elements in artistic works. Consequently, recent works have shifted towards example-guided artistic image generation. Several studies~\cite{dreambooth, shi2023instancebooth, customdiffusion, hu2022lora} developed various optimization techniques on a small collection of input images that share a common style concept. 
Inspired by Textural Inversion~(TI)~\cite{TI}, some methods~\cite{zhang2023inversion, ahn2023dreamstyler, sohn2023styledrop} propose to optimize a specific textual embedding to represent a certain style. Similarly to our work, IP-Adapter~\cite{ye2023ipadapter} trains an image adapter based on pretrained Stable Diffusion to adapt T2I models to image conditions. 
Although IP-Adapter can produce similar image variants, it fails to decouple style concepts from input images or generate images with other content through text conditions.

\vspace{-0.7em}
\subsection{Stylized Video Generation}
Building upon the foundation of stylized image generation, researchers have extended the concept to video style transfer and stylized video generation. Due to the scarcity of large-scale stylized video data, a common approach for video stylization involves applying image stylization techniques on a frame-by-frame basis. Before the advent of ML, researchers have explored methods for rendering specific artistic styles such as video watercolorization~\cite{bousseau2007video}. Early deep learning methods of video style transfer~\cite{ruder2016artistic,chen2017coherent,texler2020interactive,gao2020fast,jamrivska2019stylizing,deng2021arbitrary} apply style transfer in video sequences, generating stable stylized video sequences through the use of optical flow constraints. 
Additionally, Some video editing methods~\cite{wu2023tune,qi2023fatezero,khachatryan2023text2video,huang2023style,yang2023rerender,geyer2023tokenflow,yang2024fresco} based on pretrained T2I models also support text-guided video style transfer. Although these methods effectively improve temporal consistency, they often fail to handle frames with a large action span. Reliance on a source video also undermines flexibility. 
Similarly, certain image-to-video(I2V) methods~\cite{blattmann2023stable, xing2023dynamicrafter, xing2024tooncrafter} demonstrate capabilities in stylized video generation, particularly in the anime domain. However, I2V models still face challenges when tased with interpreting and animating highly artistic images, producing frames that veer towards realism, since real-world videos dominated its training data.

VideoComposer~\cite{wang2024videocomposer} focuses on controllable video generation, allowing multiple conditional input to govern the video generation, including structure, motion, style, etc. Although VideoComposer enables multiple controls including style, they fail to decouple style concepts, leading to limited visual quality and motion naturalness. AnimateDiff~\cite{guo2023animatediff} employs a T2I model as a base generator and adds a motion module to learn motion dynamics, which enables extending the success of personalized T2I models(e.g., LoRA~\cite{hu2022lora}, Dreambooth~\cite{dreambooth}) to video animation. However, the dependence on a personalized model restricts its ability to generate videos with arbitrary styles. Another associated research is Text2Cinemagraph~\cite{mahapatra2023text}, which utilizes pretrained text-to-image models to pioneer text-guided artistic cinemagraph creation. This approach surpasses some existing text-to-video models like VideoCrafter~\cite{chen2023videocrafter} in generating plausible motion in artistic scenes. Nevertheless, its main limitation lies in its confined applicability, primarily to landscapes, and its tendency to generate scanty motion patterns solely for fluid elements.

%---------------------------------
\vspace{-0.6em}
\section{Method}
\label{sec:method}
%---------------------------------

\begin{figure}[t]
    \centering
    \includegraphics[width=0.95\linewidth]{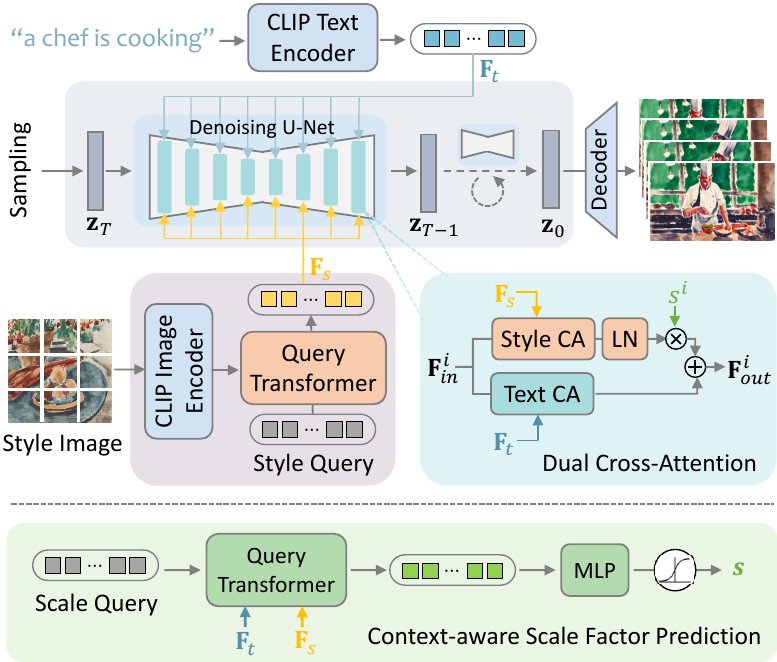}
    \vspace{-0.35cm}
    \caption{Overview of our proposed style adapter. It consists of three components, i.e. style feature extractor, dual cross-attention module, and context-aware scale factor predictor.}
    \label{fig:overview}
    \vspace{-1.4em}
\end{figure}

We propose a method to equip pre-trained Text-to-Video (T2V) models with a style adapter, allowing for the generation of stylized videos based on both a text prompt and a style reference image. The overview is illustrated in Figure~\ref{fig:overview}. In this framework, the textual description dictates the video content, while the style image governs the visual style, ensuring a disentangled control over the video generation process.
Given the limited availability of stylized videos, we employ a two-stage training strategy. Initially, we utilize an image dataset abundant in artistic styles to learn reference-based style modulation. Subsequently, adaptation finetuning on a mixed dataset of style images and realistic videos is conducted to improve the temporal quality of the generated videos.

%---------------------------------
\vspace{-0.6em}
\subsection{Reference-Based Style Modulation}
\label{subsec:style_modulation}
%---------------------------------

Our style adapter serves to extract style features from the input reference image and infuse them into the backbone features of the denoising U-Net. As mainstream T2V models~\cite{chen2023videocrafter, chen2024videocrafter2, wang2023modelscope, wang2023lavie} are generally initialized from open-source T2I Models and trained with image and video datasets in a joint strategy, they support not only text-to-video generation but also retain the capacity for text-to-image generation. To overcome the scarcity of stylized videos, we propose to train the style adapter based on a pre-trained T2V model (i.e. VideoCrafter~\cite{chen2023videocrafter}) for stylized image generation under the supervision of stylistic images.

\vspace{-0.3em}
\paragraph{Content-Style Decoupled Data Augmentation.}
\label{sec:data_aug}
We use the stylistic images from two publicly available datasets, i.e. WikiArt~\cite{phillips2011wiki} and a subset of Laion-Aesthetics~\cite{schuhmann2022laion} (aesthetics score above 6.5). In the original image-caption pairs, we observe that the captions generally contain both content and style descriptions, and some of them do not match the image content well. To promote the content-style decoupling, we use BLIP\nobreakdash-2~\cite{li2023blip2} to regenerate captions for the images and remove certain forms of style description (e.g., \textit{a painting of}) with regular expressions.
In addition, as an image contains both style and content information, it is necessary to construct a decoupling supervision strategy to guarantee the extracted style feature free of content features. Although a stylistic image may contain different local style patterns~\cite{park2019arbitrary, huo2021manifold, chen2023tssat}, we regard that a large crop of an image(e.g. 50\% of the image) still preserves a similar style representation with the full image.
% We regard that every local regions of a stylistic image share the same style representation, which not only reflects on texture and color theme but also on the structure and perceptual semantics. 
Based on this insight, we process each stylistic image to obtain the target image and style image through different strategies: for target image, we scale the shorter side of the image to 512 and then crop the target content from the central area; for style image, we scale the shorter side of the image to 800 and randomly crop a local patch with $512 \times 512$. This approach reduces the overlap between the style reference and generation target, while still preserving the global style semantics complete and consistent.

\vspace{-0.3em}
\paragraph{Style Embedding Extraction.}
CLIP~\cite{radford2021learning} has demonstrated remarkable capability in extracting visual features from open-domain images. To capitalize on this advantage, we employ a pre-trained CLIP image encoder as a feature extractor. Specifically, we utilize both the global semantic token and the full $256$ local tokens (i.e., from the final layer of the Transformer) since our desired style embedding should not only serve as an accurate style trigger for the T2V model, but also provide auxiliary feature references.
As image tokens encompass both style and content information, we further employ a trainable Query Transformer (Q-Former)~\cite{li2023blip2} to extract style embedding $\mathbf{F}_s$. We create $N$ learnable style query embeddings as input for the Q-Former, which interact with image features through self-attention layers.
%We define style as the synthesized information present in an image that characterizes its overall aesthetic, independent of content. It encompasses both high-level aesthetic semantic information, such as composition, artistic conception, and color scheme, as well as low-level texture information, including specific brush stokes, lines, and surface details.
Note that this is a commonly adopted architecture for visual condition extraction~\cite{li2023blip2, shi2023instancebooth,ye2023ipadapter, xing2023dynamicrafter}. But it is the style-content fusion mechanism that makes our proposed design novel and insightful for style modulation, as detailed below.

\begin{figure}[t]
    \centering
    \includegraphics[width=\linewidth]{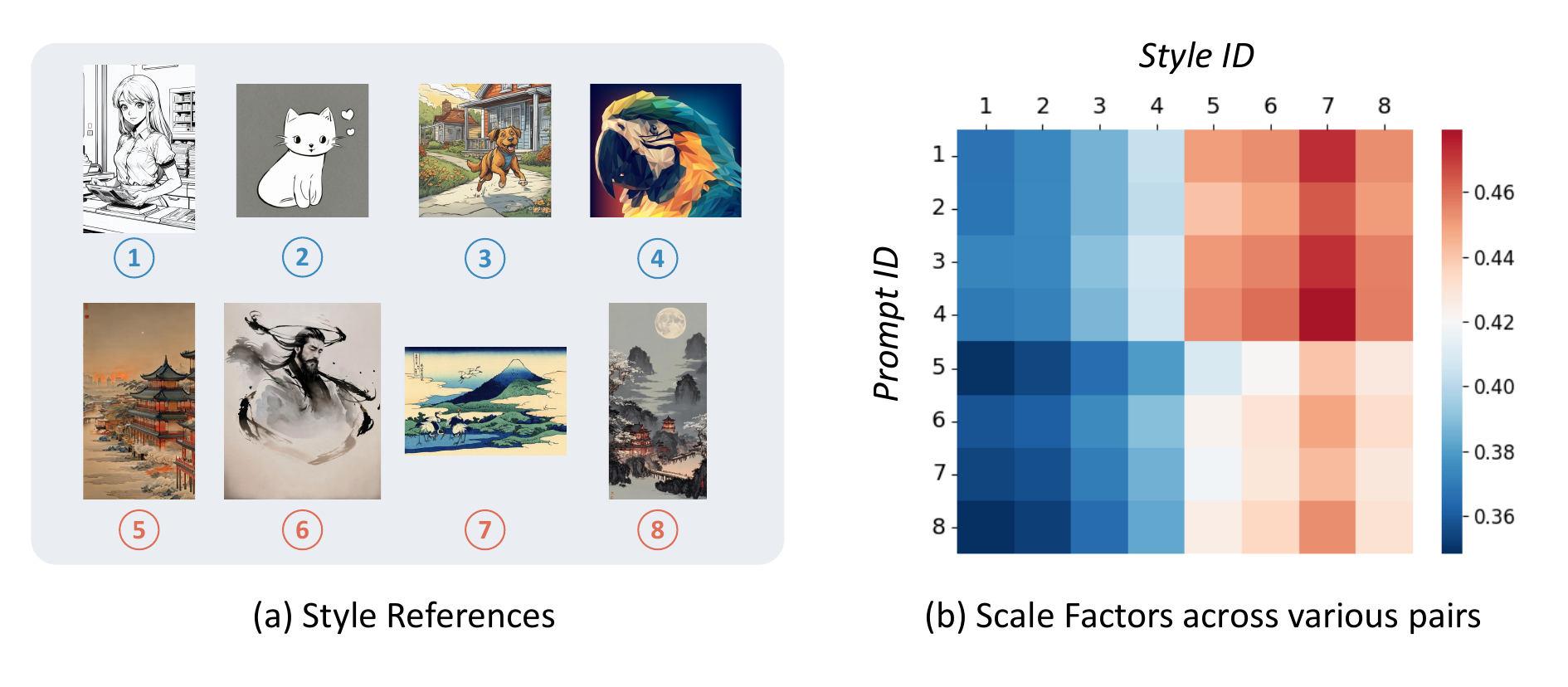}
    \vspace{-1.0cm}
    \caption{Illustration of content-style fusion scale factors across multiple input pairs. Four short prompts(less than 5 words) with prompt id $\in [1, 4]$ and four long prompts(more than 8 words) with prompt id $\in [5, 8]$ are randomly selected. Results indicate that shorter prompts and images with richer style-semantics tend to have relatively higher scale factors.} 
    \label{fig:scale_factor_viz}
    \vspace{-0.35cm}
\end{figure}

\paragraph{Adaptive Style-Content Fusion.}
\label{sec:fusion}
With the extracted style embedding, there are two ways to combine the style and text conditions, including (i) \textit{attach-to-text}~\cite{composer,gligen,ramesh2022hierarchical}: attach the style embedding to the text embedding and then interact with the backbone feature via the originally text-based cross-attention as a whole; (ii) \textit{dual cross-attention}~\cite{wei2023elite,ye2023ipadapter}: adding a new cross-attention module for the style embedding and then fuse the text-conditioned feature and style-conditioned feature.
According to our experiment (see Sec.~\ref{subsec:ablation}), solution (ii) surpasses solution (i) in disentangling the roles of text and style conditions, therefore we have adopted it as our final solution. The formula can be written as:
\vspace{-0.3em}
\begin{equation}
    \mathbf{F}_{out}^{i} = \text{TCA}(\mathbf{F}_{in}^i, \mathbf{F}_t) + s^i * \text{LN}(\text{SCA}(\mathbf{F}_{in}^i, \mathbf{F}_s)),
\end{equation}
where $\mathbf{F}_{in}^i$ denotes the backbone feature of layer $i$, LN denotes layer normalization, and TCA and SCA denote text-based cross attention and style-based cross attention respectively. $s^i$ is a scale factor learned by a context-aware scale factor prediction network, to balance the magnitudes of text-based feature and style-based feature.
The motivation is that different stylistic genres may have different emphasis on content expression. For example, the abstract styles tend to diminish the concreteness of the content, while realism styles tend to highlight the accuracy and specificity of the content. So, we propose a context-aware scale factor prediction network to predict fusion scale factors according to the input contexts.
Specifically, we create a learnable factor query, it interacts with textual features $\mathbf{F}_t$ and style features $\mathbf{F}_s$ to generate scale features via a Q-Former and then project it into layer-wise scale factors $\mathbf{s} \in \mathbb{R}^{16}$.
Figure~\ref{fig:scale_factor_viz} illustrates the learned scale factors across multiple contexts. It shows that the adaptive scale factors have a strong correlation with style genres while also depending on the text prompts. Style references with rich style-semantics(i.e., ukiyo-e style) typically yield higher scale factors to emphasize style; while complex prompts tend to produce lower scale factors to enhance content control.  This is consistent with our hypothesis to motivate our design.

%---------------------------------
\vspace{-0.5em}
\subsection{Temporal Adaptation to Stylized Features}
\label{subsec:temp_adaptation}
%---------------------------------

Given a pre-trained T2V model, the style adapter trained on image dataset works well for stylized image generation. However, it still struggles to generate satisfactory stylized videos, which is vulnerable to temporal jittering and visual artifacts.
The possible causes are that the cross-frame operations, i.e. temporal self-attention, do not involve in the process of stylized image generation, and thus induce incompatible issues. So, it is necessary to finetune the temporal self-attention with the style adapter incorporated.
Following the practice of T2V image and video joint training, the finetuning is performed on the mixed datasets of stylistic images and photorealistic videos. This is an adaptation training of temporal blocks while the other modules remain frozen, and the model converges efficiently.

\vspace{-0.5em}
\paragraph{Classifier-Free Guidance for Multiple Conditions.}
Unlike T2I models, video models exhibit a higher sensitivity to style guidance due to their limited stylized generation capabilities. Using a unified $\lambda$ for both style and context guidance may lead to undesirable generation results. Regarding this, we adopt a more flexible mechanism for multiple conditions classifier-free guidance. Building upon the vanilla text-guided classifier-free guidance, which controls context alignment by contrasting textual-conditioned distribution $\epsilon(z_t, c_t)$ with unconditional distribution $\epsilon(z_t, \varnothing)$, we introduce the style guidance with $\lambda_s$ by emphasizing the difference between the text-style-guided distribution $\epsilon(z_t, c_t, c_s)$ and the text-guided distribution $\epsilon(z_t, c_t)$. The complete formulation is as below:
\begin{equation}
    \begin{aligned}
        \hat{\epsilon}(z_t, c_t, c_s) = \epsilon(z_t, \varnothing) &+ \lambda_s(\epsilon(z_t, c_t, c_s) - \epsilon(z_t, c_t)) \\
        &+ \lambda_t(\epsilon(z_t, c_t) - \epsilon(z_t, \varnothing)),
    \end{aligned}
\end{equation}
where $c_t$ and $c_s$ denote textual and style condition respectively. $\varnothing$ denotes using no text or style conditions.
In our experiment, we follow the recommended configuration of text guidance in VideoCrafter~\cite{chen2023videocrafter}, setting $\lambda_t = 15.0$, while the style guidance is configured with $\lambda_s = 7.5$ empirically. Similarly, we set $\lambda_t = 7.5$ and $\lambda_s = 5.0$ for style-guided image generation.

%---------------------------------
\vspace{-0.3em}
\section{Experimental Results}
\label{sec:result}
%---------------------------------

%---------------------------------
\subsection{Experimental settings}
\label{subsec:experiment_setting}
%---------------------------------

\paragraph{Implementation Details.} 
We adopt the VideoCrafter~\cite{chen2023videocrafter} as our base T2V model, which shares the same spatial weights with Stable Diffusion 2.1. We first train the style modulation on image dataset, i.e. WikiArt~\cite{phillips2011wiki} and Laion-Aesthetics-6.5+~\cite{schuhmann2022laion} for 40k steps with a batch size of 32 per GPU.  In the second stage, we froze the style modulation part and only train temporal blocks of VideoCrafter, we jointly train image datasets and video datasets(subset of WebVid-10M~\cite{bain2021frozen}) for 20k steps with a batch size of 1 on video data and 16 on image data, sampling image batches with a ratio of 20\%. The training process is performed on 8 A100 GPUs and can be completed within 3 days. Furthermore, to ensure a fair comparison with some SDXL-based models~\cite{ye2023ipadapter, hertz2023style} on stylized image generation, we also trained the first stage of StyleCrafter on SDXL~\cite{podell2023sdxl}. 
%The overall training process takes about 5 days on 8$\times$V100.

% We train the style adapter on xx for xx epoches, with xx frozen and only train xxxx. In the second stage, xxx. We use xx optimizer and learning rate of xxxx. Also, training dataset.  \TODO{complete this part}

\paragraph{Testing Datasets.}
\label{sec:test_dataset}
To evaluate the effectiveness and generalizability of our method, we construct testsets comprising content prompts and style references. For content prompts, we use GPT-4~\cite{openai2023gpt4v} to generate recognizable textual descriptions from four meta-categories~(human, animal, object, and landscape). We manually filter out low-quality prompts, retaining 20 image prompts and 12 video prompts. For style references, we collect 20 stylized images and 8 sets of style images with multi-reference~(each contains 5 to 7 images in similar styles) from the Internet. In total, the test set contains 400 pairs for stylized image generation, and 300 pairs for stylized video generation (240 single-reference pairs and 60 multi-reference pairs). Details are available in the supplementary materials.

% 1. Content Prompt(generated from GPT4)
% 2. Style Image(selected from internet)
% 3. Content Image(generated from SD, for some style transfer method)
% 4. Style Prompt(generated from GPT4-v, for sd and video-crafter)

\paragraph{Evaluation Metrics.}
\label{sec:eval_metrics}
Following previous practice~\cite{zhang2023inversion, sohn2023styledrop, wang2023styleadapter}, we employ CLIP-based~\cite{radford2021learning} scores and DINO-based~\cite{caron2021emerging} scores to measure the text alignment and style conformity. Following EvalCrafter~\cite{liu2023evalcrafter}, we measure the temporal consistency of video generation by (i) calculating clip scores between contiguous frames and (ii) calculating the warping error on every two frames with estimated optical flow. 
Note that these metrics are not perfect. For example, one can easily achieve a close-to-1 style score by entirely replicating the style reference. Similarly, stylized results may yield inferior text scores compared to realistic results, even though both accurately represent the content descriptions. We recommend a comprehensive consideration of both CLIP-based text scores and style scores, rather than relying solely on a single metric.

\paragraph{User Preference Study.}
In addition to quantitative analysis, we conducted a user study to make comparisons among our method, VideoCrafter, Gen-2, and AnimateDiff in the context of single-reference and multi-reference stylized video generation. Users are instructed to select their preferred option based on style conformity, temporal quality, and all options fulfill text alignment for each comparison pair. We randomly chose 15 single-reference pairs and 10 multi-reference pairs, collecting 1125 votes from 15 users. 
Further details can be found in the supplementary materials.

\begin{table*}[!t]
\centering
\caption{Quantitative comparison on single-reference style-guided T2I generation. We conduct evaluation on a test set of 400 pairs. \textbf{Bold}: Best. }
\label{tab:img_quan_clip}
\vspace{-1em}
\resizebox{0.93\linewidth}{!}{
  \begin{tabular}{ccccccccc} % {@{}lc@{}}
    \toprule
     \multirow{2}{*}{Method} & \multicolumn{4}{c}{\textbf{Stable Diffusion 2.1 based}} & \multicolumn{4}{c}{\textbf{SDXL based}} \\
    \cmidrule(lr){2-5}\cmidrule(lr){6-9}
     & Dreambooth & InST & SD* & Ours & IP-Adapter-Plus & Style-Aligned & SDXL* & Ours(SDXL)  \\
    \midrule
    \texttt{CLIP-Text} $\uparrow$ & \textbf{0.3047}  & 0.3004 & 0.2766 & 0.3028 & 0.2768 & 0.2254 & 0.2835 & \textbf{0.2918} \\
    \texttt{CLIP-Style} $\uparrow$ & 0.3459 & 0.3708 & 0.4183 & \textbf{0.4836} & 0.5182 & 0.5515 & 0.4348 & \textbf{0.5615} \\
    \texttt{DINO-Style} $\uparrow$ & 0.2278 & 0.2587 & 0.2890 & \textbf{0.3652} & 0.4367 & 0.4395 & 0.2912 & \textbf{0.4514} \\
    \bottomrule
  \end{tabular}
}
\end{table*}

%---------------------------------
\subsection{Style-Guided Text-to-Image Generation}
\label{subsec:image_eval}
%---------------------------------

As mentioned in Sec.~\ref{subsec:style_modulation} and Sec.~\ref{subsec:experiment_setting}, our proposed method also supports to generate stylized images~(using model before temporal finetuning). We are interested to evaluate our method against state-of-the-art style-guided T2I synthesis methods, which are better-established than video counterparts. The competitors include optimization-based methods like DreamBooth~\cite{dreambooth}, inversion-based methods such as InST~\cite{zhang2023inversion} and Style-Aligned~\cite{hertz2023style}, and adapter-based methods like IP-Adapter-Plus~\cite{ye2023ipadapter}. Besides, we consider two unique competitors: SD*~\cite{ldm} and SDXL*~\cite{sdxl}~(text-to-image models equipped with GPT-4V~\cite{openai2023gpt4v}, where GPT-4V generates textual descriptions about the reference's style and merges them with content prompts as input for models). This comparison aims to validate the advantages of employing image conditions to enhance stylized generation instead of relying solely on text conditions.
Implementation details of competitors are available in supplementary materials.
% For each style, DreamBooth and CustomDiffusion are optimized with the provided single reference image to learn the customized concept of style.
%This setting may degrade their performance (because they usually requires 5~20 images to learn a concept) but it is fair comparison since all the methods are only provided with a single style image.

The quantitative comparison is tabulated in Table~\ref{tab:img_quan_clip}. 
% Although our method achieved the best performance on both metrics, we would like to emphasise that clip-based metrics are imperfect.As discussed in Sec.~\ref{sec:eval_metrics}, the CLIP-Text is measured by the similarity between content text embedding and stylized image embedding, the stylistic appearance actually hinders the metric in some extent, which makes those methods with weak stylistic effects (i.e. close to photorealism) achieve superior scores. 
Results reveal that Dreambooth~\cite{dreambooth} and InST~\cite{zhang2023inversion} struggle to accurately capture the style from various style references and exhibit low style conformity. SD*~\cite{ldm} and SDXL~\cite{sdxl} demonstrate good stylistic ability but still fail to reproduce the style of the reference image, possibly because of the text’s inherent clumsiness in expressing specific styles despite utilizing the powerful GPT4V for visual style understanding. IP-Adapter~\cite{ye2023ipadapter} and style-aligned~\cite{hertz2023style} generate aesthetically pleasing images, while their style-content decoupled learning is not perfect and exhibits limited control over content.
In contrast, our method efficiently generates high-quality stylized images that align with the content of the texts and resemble the style of the reference image. Our method demonstrates stable stylized generation capabilities when dealing with various types of prompts.

%---------------------------------
\subsection{Style-Guided Text-to-Video Generation}
\label{subsec:video_eval}
%---------------------------------

\begin{figure*}[t!]
    \centering
    \includegraphics[width=0.87\linewidth]{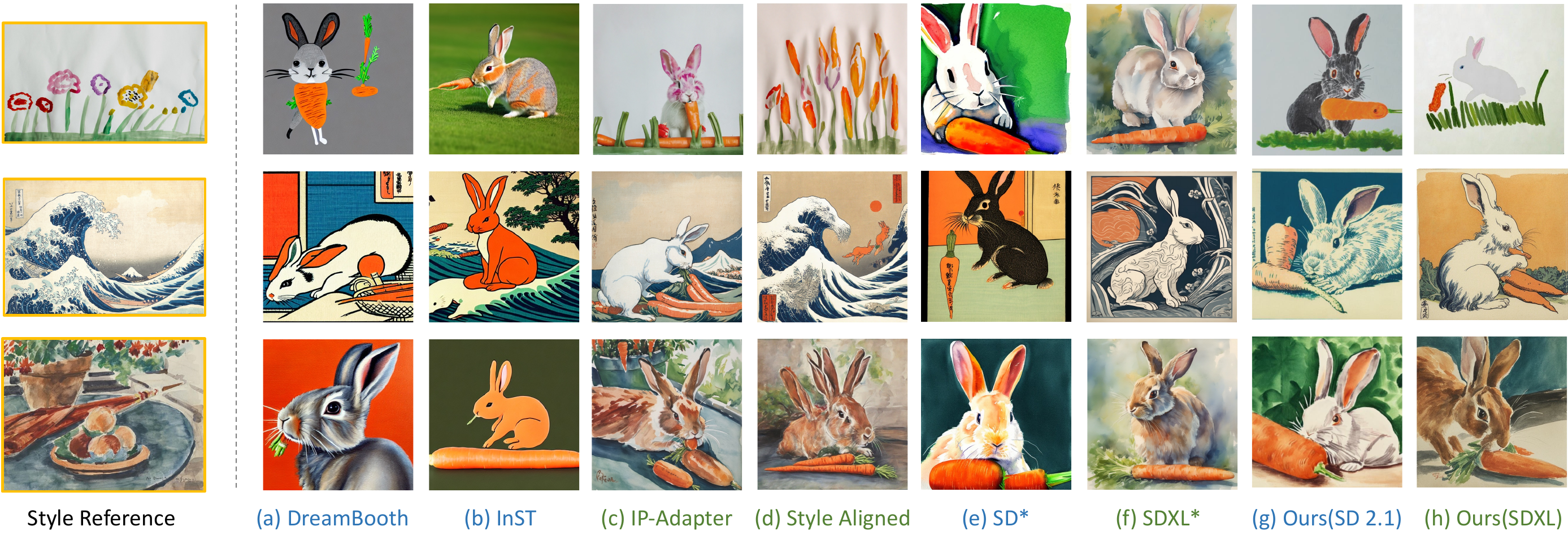}
    \vspace{-1em}
    % \vspace{-2em}
    \caption{Visual comparison on style-guided T2I generation. \textcolor[RGB]{46, 117, 182}{Blue}: methods based on SD 2.1. \textcolor[RGB]{84, 130, 53}{Green}: based on SDXL. Prompt: A rabbit nibbling on a carrot.} 
    \label{fig:result_img}
    \vspace{-1em}
\end{figure*}
\begin{figure*}[!h]
    \centering
    \includegraphics[width=0.9\linewidth]{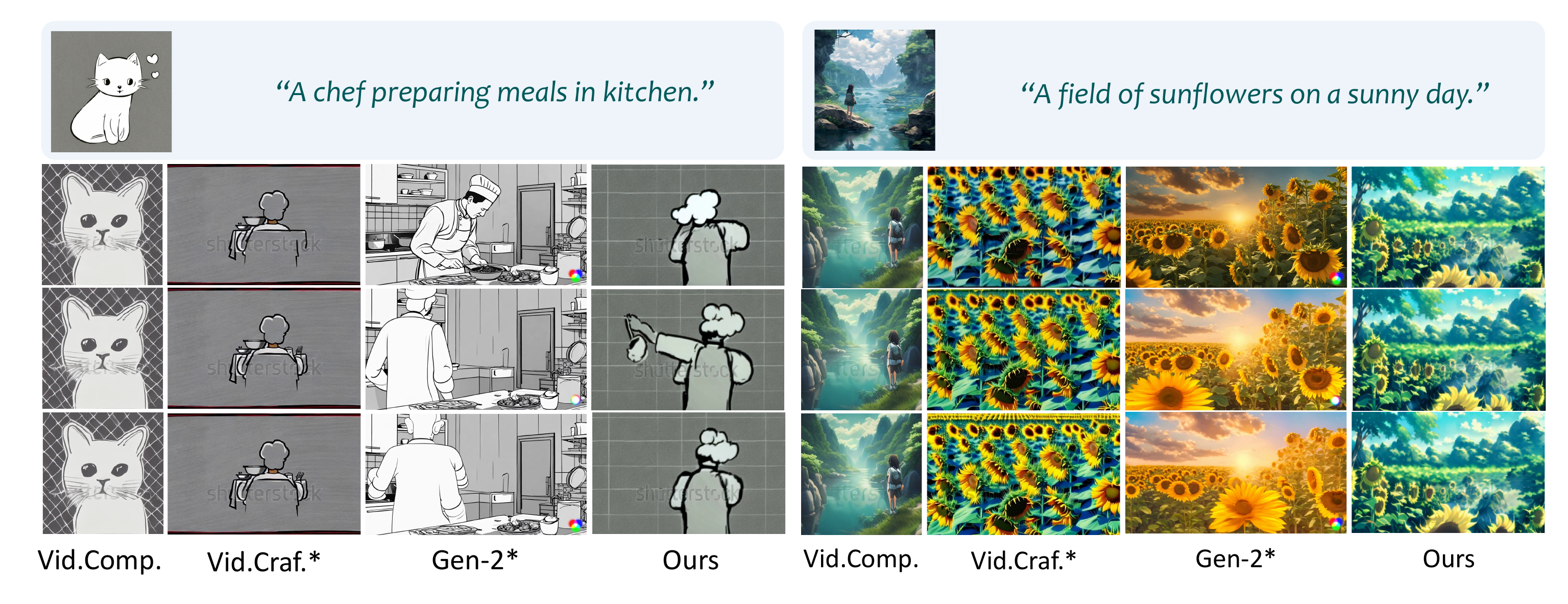}\vspace{-1.5em}
    \caption{Visual comparison of single-reference guided T2V generation. Vid.Comp.: VideoComposer, Vid.Craf.: VideoCrafter
    %Our approach accurately captures the style of the reference image and generates videos that satisfy both text alignment and style conformity.
    } 
    \label{fig:result_video}
    \vspace{-1em}
\end{figure*}
\begin{figure*}[!h]
    \centering
    \includegraphics[width=0.87\linewidth]{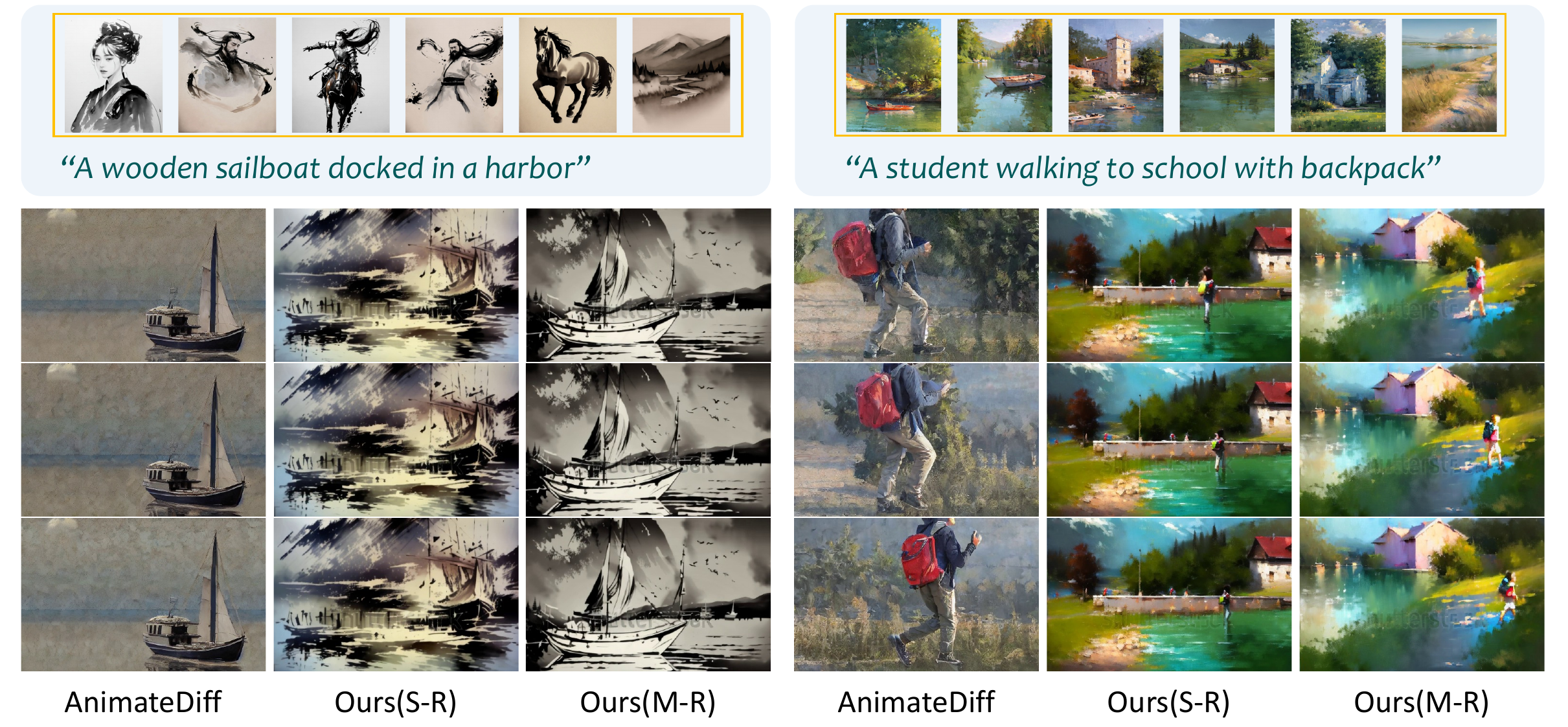}
    \vspace{-1em}
    \caption{Qualitative comparison of multi-reference style-guided T2V generation. S-R: Single-Reference, M-R: Multi-Reference
    %AnimateDiff tends to generate close-to-reamlism results despite the style references are typical artistic styles. In contrast, our approach performs better in text alignment, style conformity and temporal consistency.
    } 
    \vspace{-1em}
    \label{fig:result_multi_ref}
\end{figure*}

\begin{table}[!t]
\centering
\caption{Quantitative comparison of style-guided T2V generation. Top 3 rows: single-reference based gudiance. Bottom 3 rows: multi-reference based guidance. S-R: Single-Refernce, M-R: Multi-Reference, W.E.: Warping Errors.}
\label{tab:video_quan_clip}
\vspace{-1em}
\resizebox{\linewidth}{!}{
% \scriptsize
% \renewcommand\arraystretch{1.2}
\begin{tabular}{c c c cc} % {@{}lc@{}}
    % \toprule
    \toprule
    \multirow{2}{*}{Methods} & \multirow{2}{*}{CLIP-Text $\uparrow$} & \multirow{2}{*}{CLIP-Style $\uparrow$} & \multicolumn{2}{c}{Temporal Consistency} \\
    \cmidrule(lr){4-5}
      & & &
     ~CLIP-Temp $\uparrow$~ & \makecell{W.E.($\times10^{-3}$) $\downarrow$}\\
    \midrule
    VideoComposer & 0.0468 & \textbf{0.7306} & 0.9853 & \textbf{9.903} \\
    VideoCrafter* & 0.2209 & 0.3124 & 0.9757 & 61.41 \\
    Ours & \textbf{0.2726} & 0.4531 & \textbf{0.9892} & 18.73 \\
    \midrule
    ~AnimateDiff & \textbf{0.2867} & 0.3528 & 0.8903 & 37.17 \\
    Ours(S-R) & 0.2661 & 0.4803 & 0.9851 & 14.13 \\
    Ours(M-R) & 0.2634 & \textbf{0.4887} &\textbf{0.9852} & \textbf{9.396} \\
    \bottomrule
  \end{tabular}
}
\end{table}

\begin{table}[!t]
\centering

\caption{User study statistics of the selection rate for text alignment(\texttt{Text}), and preference rate for style conformity(\texttt{Style}) and temporal quality(\texttt{Temporal}). Top 3 rows: single-reference based guidance. Bottom 2 rows: multi-reference based guidance.}
\label{tab:video_quan_multi_ref}
\vspace{-1em}

\small
\begin{tabular}{c c c c} % {@{}lc@{}}
    % \toprule
    \toprule
    Methods & \texttt{Text} $\uparrow$ & \texttt{Style} $\uparrow$ & \texttt{Temporal} $\uparrow$ \\
    \midrule
    VideoCrafter* & 0.391 & 8.0\% & 4.4\%  \\
    Gen-2* & 0.747 & 23.1\% & 51.1\% \\
    Ours & 0.844 & 68.9\% & 44.4\% \\
    \midrule
    ~AnimateDiff~ & 0.647 & 10.0\% & 19.3\% \\
    Ours(M-R) & 0.907 & 90.0\% & 80.7\% \\
    \bottomrule
\end{tabular}
\end{table}

Existing approaches for style-guided video generation can be divided into two categories: one is the single-reference based methods that are usually tuning-free, e.g. VideoComposer~\cite{wang2024videocomposer}; the other is the multi-reference based methods that generally requires multiple references for fine-tuning, e.g. AnimateDiff~\cite{guo2023animatediff}. We make comparisons with these methods, respectively. 
% Apart from the quality metrics, we further conduct user study to evaluate the stylized video results, including text alignment, style conformity and temporal quality.

\vspace{-0.5em}
\paragraph{Single-Reference based Guidance.}VideoComposer~\cite{wang2024videocomposer} is a controllable video generation model that allows multiple conditional inputs including style reference image. It is a natural competitor of our method. 
Besides, we construct two additional comparative methods, i.e. VideoCrafter* and Gen2*, which extend VideoCrafter~\cite{chen2023videocrafter} and Gen2~\cite{Gen-2}, the state-of-the-art T2V models in open-source and close-source channels respectively, to make use of style reference images by utilizing GPT-4V~\cite{openai2023gpt4v} to generate style prompts from them.
% The evaluation is conducted on 240 text-style pairs, as introduced in Sec.~\ref{sec:test_dataset}. 
The quantitative comparison is tabulated in Table~\ref{tab:video_quan_clip}. Several typical visual examples are illustrated in Figure~\ref{fig:result_video}.

We can observe that: 
(i) VideoComposer tends to copy content from style references and struggles to generate text-aligned content, which is possibly because of the invalid decoupling learning. Consequently, its results exhibit abnormally high style conformity and very low text alignment. In addition, VideoComposer often generates static videos, thus having the lowest warping errors, but this does not mean that their results perform best in temporal quality. 
(ii) VideoCrafter* exhibits limited stylized generation capabilities, producing videos with diminished style and disjointed movements. Gen-2* demonstrates superior stylized generation capabilities. However, Gen-2 is still limited by the inadequate representation of style in textual descriptions, and is more prone to sudden changes in color and luminance. (iii) In comparison, our method captures styles more effectively and reproduces them in the generated results.

\begin{figure}[!t]
    \centering
    \includegraphics[width=\linewidth]{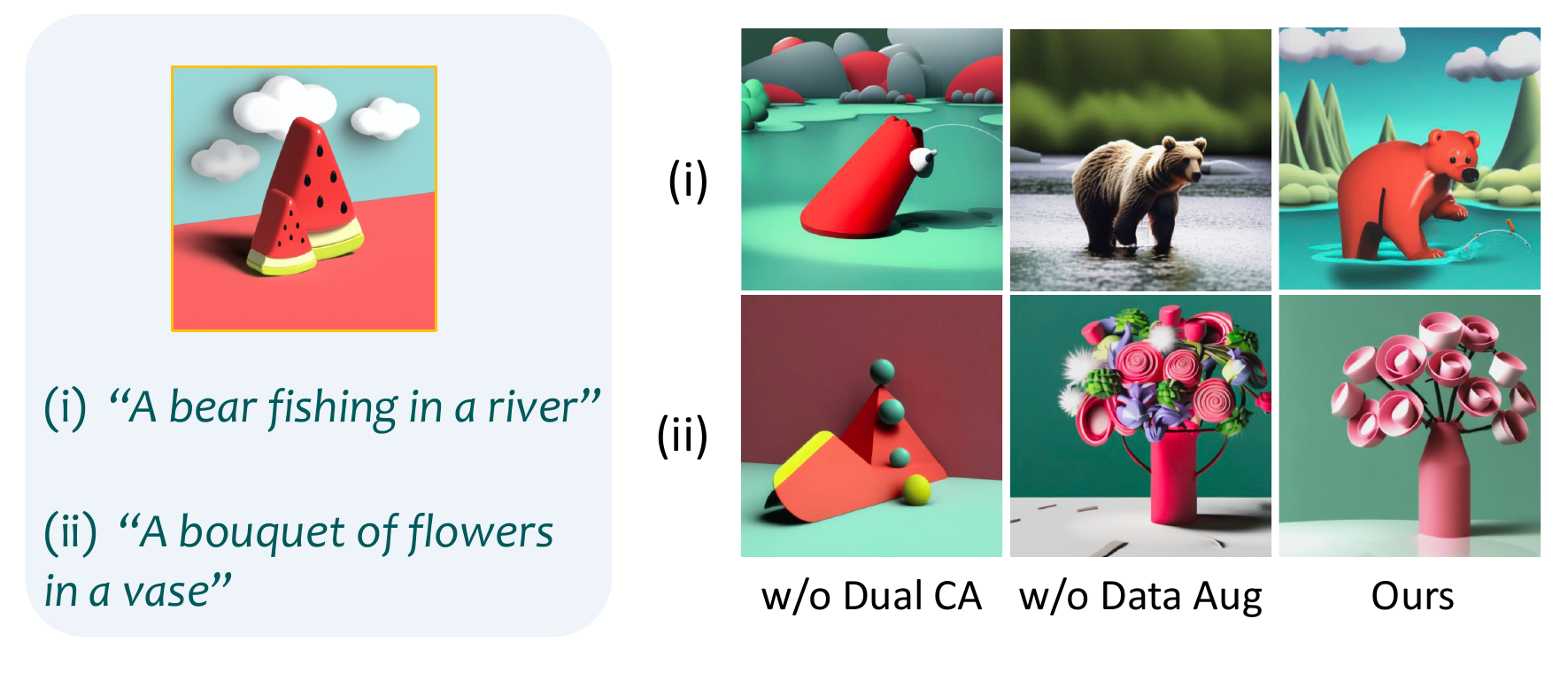}
    \vspace{-3em}
    \caption{Effects of dual cross-attention and data augmentation.} 
    \label{fig:ablation_img_1_2}
    \vspace{-1em}
\end{figure}

\vspace{-0.5em}
\paragraph{Multi-Reference based Guidance.}
AnimateDiff~\cite{guo2023animatediff} denotes a paradigm to turn personalized SD (i.e., SD fine-tuned on specific-domain images via LoRA~\cite{hu2022lora} or Dreambooth~\cite{dreambooth}) for video generation, namely combined with pre-trained temporal blocks of T2V models. It can generate very impressive results if the personalized SD is carefully prepared, however, we find it struggle to achieve as satisfactory results if only a handful of style reference images are available for training.
We conduct an evaluation on 60 text-style pairs with multi-references, as presented in Sec.~\ref{sec:test_dataset}. We train Dreambooth~\cite{dreambooth} models for each style and incorporate them into AnimateDiff based on their released codebase. Thanks to the flexibility of Q-Former, our method also supports multiple reference images in a tuning-free fashion, i.e. computing the image embeddings of each reference image and concatenating all embeddings as input to the Q-Former.
Results are compared in Table~\ref{tab:video_quan_multi_ref} and Figure~\ref{fig:result_multi_ref} respectively.

According to the results, AnimateDiff struggles to achieve high-fidelity stylistic appearance while tends to generate close-to-realism results despite the style references are typical artistic styles. In addition, it is vulnerable to temporal artifacts. As the trained personalized-SD can generate decent stylistic images (provided in the supplementary materials), we conjecture that the performance degradation is caused by the incompatibility from the pre-trained temporal blocks and independently trained personalized-SD models, which not only interrupts temporal consistency but also weakens the stylistic effect.
In contrast, our method can generate temporal consistent videos with high style conformity to the reference images and accurate content alignment with the text prompts. Furthermore, using multiple references can further promote the performance, which offers additional advantages in practical applications.

%---------------------------------
\subsection{Ablation Study}
\label{subsec:ablation}
%---------------------------------

\begin{figure}[!t]
    \centering
    \includegraphics[width=\linewidth]{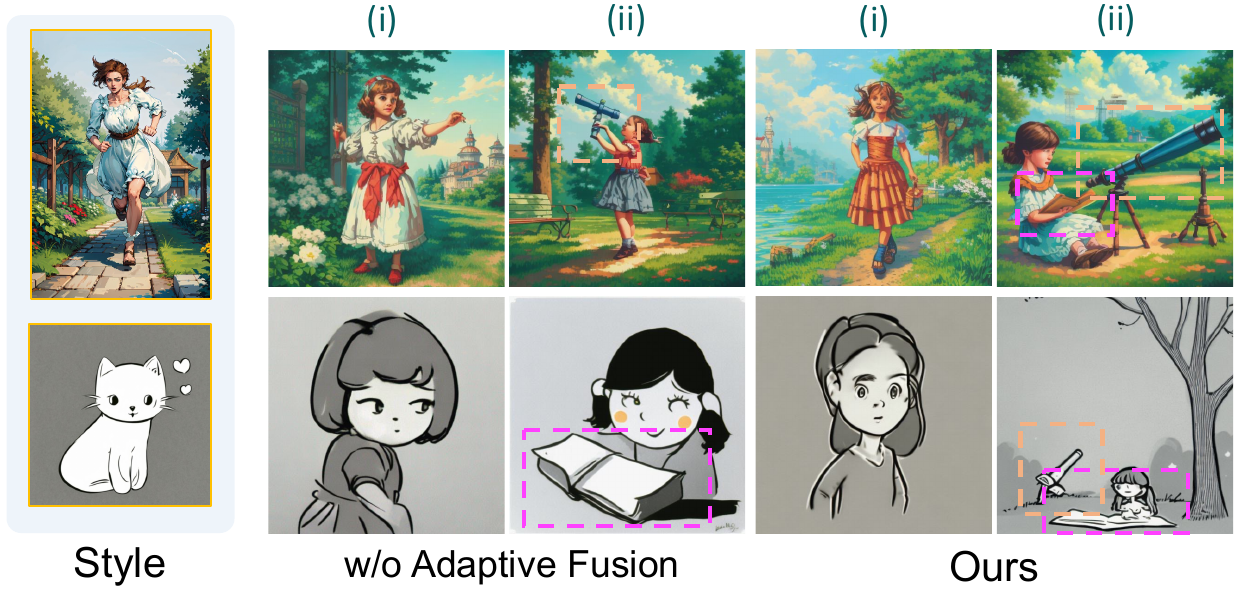
    }
    \vspace{-2em}
    \caption{Effect of adaptive content-style fusion. It shows superiority in generalization to extreme cases, e.g. long text description. Two text prompts are used: (i) A little girl; (ii) \textit{A little girl \textcolor[RGB]{255, 64, 255}{reading a book} in the park, with \textcolor[RGB]{244, 177, 131}{a telescope} nearby pointed at the sky}.}
    \label{fig:ablation_no_fusion}
    \vspace{-0.7em}
\end{figure}

% We make ablation studies on some important designs of our method, including data augmentation, module architectures, and training strategies, to validate their effectiveness.

\paragraph{Data Augmentation.}
We first study the effectiveness of content-style decoupled data augmentation. As depicted in Table~\ref{tab:ablation_img}, training with the original image-caption pairs restricts the model's ability to extract style representations, leading to lower style conformity. For example, as shown in Figure~\ref{fig:ablation_img_1_2}, method without data augmentation fails to capture the "3D render" style from the reference.

\paragraph{Dual Cross-Attention.} 
As discussed in Sec.~\ref{subsec:style_modulation},  we make a comparison between \textit{attach-to-text} and \textbf{dual cross-attention} to study their effects. Results are presented in Table ~\ref{tab:ablation_img} and Figure~\ref{fig:ablation_img_1_2}, revealing that \textit{attach-to-text} tends to directly fuse the content from the reference image and the text prompts rather than combining the text-based content and image-based style. This indicates the effectiveness of \textbf{dual cross-attention} in facilitating content-style decoupling.

\paragraph{Adaptive Style-Content Fusion.}
As previously discussed in Figure~\ref{fig:scale_factor_viz}, our proposed adaptive style-content fusion module demonstrates effectiveness in adaptively processing various conditional contexts. It benefits the generalization ability of model to deal with diverse combinations of content prompt and style image. Figure~\ref{fig:ablation_no_fusion} reveals that although the baseline can handle easy prompt inputs like "A little girl", it struggles to accurately generate all objects described in longer prompts. In contrast, the adaptive fusion module can achieve decent text alignment for long text descriptions thanks to its flexibility to adaptive balance between content and style.

\paragraph{Two-Stage Training Scheme.}
Our proposed training scheme consists of two stages, i.e., style adapter training and temporal adaption. To show its necessity, we build two baselines: (i) \textit{w/o Temporal Adaption}: that we train a style adapter on image data and apply it directly to stylized video generation without finetuning; (ii) \textit{joint training}: that we conduct style adapter training and temporal blocks finetuning on image-video dataset simultaneously.
As depicted in Table~\ref{tab:ablation_video}, baseline (i) exhibits inferior temporal consistency when applied directly to video, and undermines the content alignment and style conformity. As for baseline (ii), the learning of style embedding extraction seems to be interfered by the joint finetuning of temporal blocks, which impedes it to generate desirable stylized videos.

\begin{table}[!t]
\centering

\caption{Ablation studies on style modulation designs. The performance is evaluated based on the style-guided T2I generation.}
\label{tab:ablation_img}
\vspace{-1em}
    
\small
  \begin{tabular}{ccc} % {@{}lc@{}}
    % \toprule
    \hline
    Methods & \texttt{CLIP-Text} $\uparrow$ & \texttt{CLIP-Style} $\uparrow$ \\
    \hline
    Ours & 0.3028 & 0.4836 \\
    ~w/o Data Augmentation & 0.3173 & 0.4005 \\
    w/o Dual Cross Attention & 0.0983 & 0.7332 \\
    w/o Adaptive Fusion & 0.2807 & 0.4925 \\
    \hline
  \end{tabular}
\vspace{-1em}
\end{table}

\begin{table}[!t]
\centering
\vspace{1em}

\caption{Ablation study on our two-stage training scheme.}
\vspace{-1em}
\label{tab:ablation_video}
\resizebox{\linewidth}{!}{

\begin{tabular}{c c c cc} % {@{}lc@{}}
    % \toprule
    \toprule
    \multirow{2}{*}{Methods} & \multirow{2}{*}{CLIP-Text $\uparrow$} & \multirow{2}{*}{CLIP-Style $\uparrow$} & \multicolumn{2}{c}{Temporal Consistency} \\
    \cmidrule(lr){4-5}
      & & &
     ~CLIP-Temp $\uparrow$~ & \makecell{W.E.($\times10^{-3}$) $\downarrow$}\\
    \midrule
    w/o Temporal Adaption  & 0.2691 & 0.3923 & 0.9612 & 47.88 \\
    Joint Training & 0.3138 & 0.2226 & 0.9741 & 24.74 \\
    Two-Stage(ours) & \textbf{0.2726} & \textbf{0.4531} & \textbf{0.9892} & \textbf{18.73} \\
    \bottomrule
  \end{tabular}
}

    \vspace{-1em}
  
\end{table}

%
% \begin{figure}[!t]
%     \centering
%     \includegraphics[width=\linewidth]{figures/ablation_two_stage.pdf}
%     \vspace{-2em}
%     \caption{Comparison on the effects of different training schemes.}
%     \label{fig:ablation_two_stage}
%     \vspace{-1em}
% \end{figure}

%-------------------------------
\section{Conclusion and Limitations}
\label{sec:conclusion}
%-------------------------------

We have presented StyleCrafter, a generic method enabling pre-trained T2V model for video generation in any style by providing a reference image. To achieve this, we made exploration in three aspects, including the architecture of style adapter, the content and style feature fusion mechanism, and some tailor-made strategies for data augmentation and training stylized video generation without stylistic video data.  
All of these components allow our method to generate high quality stylized videos that align with text prompts and conform to style references.
Extensive experiments have evidenced the effectiveness of our proposed designs and comparisons with existing competitors demonstrate the superiority of our method in visual quality and efficiency.
Anyway, our method also has certain limitations, e.g., unable to generate desirable results when the reference image can not represent the target style sufficiently or the presented style is extremely unseen. Further explorations are demanded to address those issues.
%Due to the absence of high-quality stylized video data for training, the generation results may exhibit degraded motion under certain styles, e.g. abstract artwork. And the image quality of the generation video is somewhat diminished in comparison to the stylized image generation. We have further explored the potential of incorporating a small set of stylized video data to address these limitations in Supp.

\section*{ACKNOWLEDGMENTS}
This work was partly supported by the National Natural Science Foundation of China  (Grant No. 61991451) and the Shenzhen Science and Technology Program (JSGG20220831093004008).

\citestyle{acmauthoryear}
\bibliographystyle{ACM-Reference-Format}
\bibliography{reference}

\clearpage
% \clearpage
\setcounter{page}{1}
\setcounter{section}{0}
\setcounter{figure}{0}
\setcounter{table}{0}
% \maketitlesupplementary

\renewcommand{\thesection}{\Alph{section}}
\renewcommand{\thetable}{S\arabic{table}}
\renewcommand{\thefigure}{S\arabic{figure}}

%%%%%%%%%%%%%%%%%%%%%%%%%%%%%%%%%
% \renewcommand{\contentsname}{}
% \tableofcontents

\noindent Our Supplementary Material consists of 7 sections:

\begin{itemize}[leftmargin=2.0em]
    \item Section~\ref{sec:supp_implement} provides a detailed statement of our experiments, including the implementation details of comparison methods, and details of our test set.
    \item Section~\ref{sec:supp_eval} provides a detailed statement of our evaluation, including the details of evaluation metrics, and details of the user study.
    \item Section~\ref{sec:supp_extend_comp} adds more comparison experiments, including the comparison with StyleDrop, comparison in multi-reference stylized image generation, and comparison with style transfer methods.
    \item Section~\ref{sec:supp_extend_ablation} adds additional ablation study on different adapter architecture.
    \item Section~\ref{sec:supp_app} explores the extended application of StyleCrafter, including the collaboration with depth control.
    \item Section~\ref{sec:supp_more_result} demonstrates more results of our methods.
    \item Section~\ref{sec:supp_limitation} discusses the limitations.
\end{itemize}

% In addition to this supplementary document, we highly encourage readers or reviewers to explore our \textbf{local HTML webpage for video comparisons}. This resource provides a more interactive and visual insight, facilitating in-depth comprehension about StyleCrafter's advantages and limitations.

%%%%%%%%%%%%%%%%%%%%%%%%%%%%%%%%%
\section{Implementation Details} 
\label{sec:supp_implement}

\subsection{Comparison methods}

For all comparison methods, we follow the instructions from the official papers and open-source implementations. Since some methods including Dreambooth and InST require additional finetuning, we provide all implementation details as follows:

\paragraph{Dreambooth} \label{sec:supp_train_dreambooth} Dreambooth~\cite{dreambooth} aims to generate images of a specific concept (e.g., style) by finetuning the entire text-to-image model on one or serveral images. We train Dreambooth based on Stable Diffusion 1.5. The training prompts are obtained from BLIP-2~\cite{li2023blip2}, and we manually add a style postfix using the rare token "sks". For example, "two slices of watermelon on a red surface in sks style" is used for the first style reference in Table~\ref{tab:supp_style_ref}. We train the model for 500 steps for single-reference styles and 1500 steps for multi-reference styles, with learning rates of $5 \times 10^{-6}$ and a batch size of $1$. The training steps are carefully selected to achieve the balance between text alignment and style conformity. 

% \paragraph{CustomDiffusion} CustomDiffusion~\cite{customdiffusion} propose an efficient method for fast tuning text-to-image models for certain styles or concepts. We train CustomDiffusion based on Stable Diffusion 1.5. Similar to Dreambooth, we obtained training prompts from BLIP-2~\cite{li2023blip} and we manually add postfix like "in <new1> style". We generate a set of 200 regularization images from mannually designed instant prompts for each style. We train the model for 500 steps for single-reference styles and 1500 steps for multi-reference styles, with learning rates of $1 \times 10^{-5}$ and a batch size of $2$.

\paragraph{InST} InST~\cite{zhang2023inversion} propose a inversion-based method to achieve style-guided text-to-image generation through learning a textual description from style reference. We train InST for 1000 steps with learning rates of $1 \times 10^{-4}$ and a batch size of $1$.

\paragraph{StableDiffusion 2.1 and SDXL} We extend Stable Diffusion to style-guided text-to-video gerneration by utilizing GPT-4v to generate style descriptions from style reference. Details about style descriptions can be found in Table~\ref{tab:supp_style_ref}

\paragraph{IP-Adapter} IP-Adapter~\cite{ye2023ipadapter} propose to train an image-conditioned adapter to generate images from image prompts. We use the official checkpoint of IP-Adapter-Plus(SDXL) for evaluation. Note that IP-Adapter is primarily designed for image variants and other editing tasks. When conducted with its default scale value $s=1$, IP-Adapter tends to simply reconstruct style references, which actually underestimates the ability of IP Adapters in stylized generation. During the evaluation, \textbf{we adjust the scale value to 0.5 to ensure a more balanced comparison}.

\paragraph{Style Aligned} Style Aligned~\cite{hertz2023style} design a self-attention sharing mechanism to ensure constant style among different samples, supporting both stylized generation and style transfer tasks. We conduct the official implementation on SDXL during the evaluation.

\paragraph{VideoCrafter and Gen-2} Similar to SD*, We use VideoCrafter~\cite{chen2023videocrafter} $320\times512$ Text2Video Model and Gen-2~\cite{Gen-2} equipped with GPT-4v to generate stylized videos from style references and text prompts.

\paragraph{AnimateDiff} AnimateDiff~\cite{guo2023animatediff} aims to extend personalized T2I model(i.e., Dreambooth or LoRA~\cite{hu2022lora}) for video generation. To compare with AnimateDiff, we first train personalized dreambooth models for each group of multi-reference style images, then we incorporate them into AnimateDiff based on their released codebase. We did not use LoRA because we observed that AnimateDiff fails to turn LoRA-SD for video generation in most cases.

\subsection{Testing Datasets}

We provide a detailed description of the testing datasets. 

\begin{figure*}[!h]
    \centering
    \includegraphics[width=0.85\linewidth]{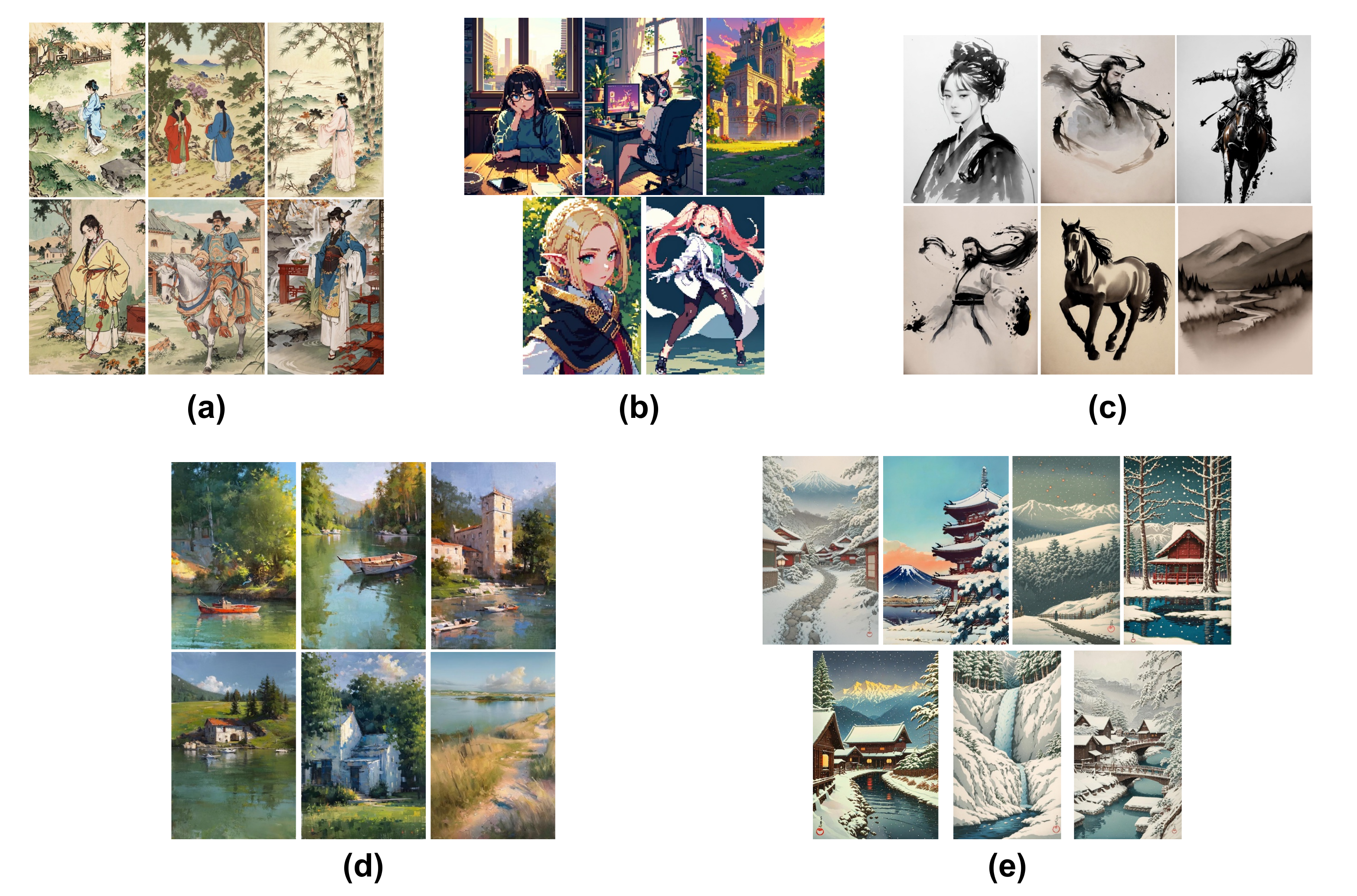}
    \vspace{-1.2em}
    \captionof{figure}{Multiple references in the testset} 
    \label{fig:supp_multi_ref}
    \vspace{0.3em}

    \centering
    % \vspace{-0.5em}
    \captionof{table}{Text prompts used in the testset for image generation}
    \label{tab:supp_text_prompt_img}
    \vspace{-2.4em}
    \renewcommand{\arraystretch}{1.3}
    %\resizebox{0.99\linewidth}{!}{%
    \small
    \vspace{1.5em}
    \begin{tabular}{>{\centering\arraybackslash}p{5.5cm}>{\centering\arraybackslash}p{1.9cm}|>{\centering\arraybackslash}p{6.3cm}>{\centering\arraybackslash}p{1.9cm}} 
        %\toprule
        \hline
        Prompt & Meta Category & Prompt & Meta Category \\
        \hline
        A man playing the guitar on a city street. & Human & A flock of birds flying gracefully in the sky. &  Animal \\
        A woman reading a book in a park. & Human & A colorful butterfly resting on a flower. &  Animal \\
        A couple dancing gracefully together. & Human & A bear fishing in a river. &  Animal \\
        A person sitting on a bench, feeding birds. & Human & A dog running in front of a house. &  Animal \\
        A person jogging along a scenic trail. & Human & A rabbit nibbling on a carrot. &  Animal \\
        A bouquet of flowers in a vase. & Object & A cobblestone street lined with shops and cafes. &  Landscape \\
        A telescope pointed at the stars. & Object & A modern cityscape with towering skyscrapers. &  Landscape \\
        A rowboat docked on a peaceful lake. & Object & A winding path through a tranquil garden. &  Landscape \\
        A lighthouse standing tall on a rocky coast. & Object & An ancient temple surrounded by lush vegetation. &  Landscape \\
        A rustic windmill in a field. & Object & A serene mountain landscape with a river flowing through it. &  Landscape \\
        \hline
    \end{tabular}
    %}

    %%%%%%%%%%%%%%%%%%%%%%%%%
    \vspace{1.5em}
    \centering
    \captionof{table}{Text prompts used in the testset for video generation}
    \label{tab:supp_text_prompt_vid}
    \vspace{-1em}
    \begin{tabular}{>{\centering\arraybackslash}p{5.5cm}>{\centering\arraybackslash}p{1.9cm}|>{\centering\arraybackslash}p{6.3cm}>{\centering\arraybackslash}p{1.9cm}} 
        %\toprule
        \hline
        Prompt & Meta Category & Prompt & Meta Category \\
        \hline
        A street performer playing the guitar. & Human & A bear catching fish in a river. &  Animal \\
        A chef preparing meals in kitchen. & Human & A knight riding a horse through a field. &  Animal \\
        A student walking to school with backpack. & Human & A wolf walking stealthily through the forest. &  Animal \\
        A campfire surrounded by tents. & Object & A river flowing gently under a bridge. &  Landscape \\
        A hot air balloon floating in the sky. & Object & A field of sunflowers on a sunny day. &  Landscape \\
        A rocketship heading towards the moon. & Object & A wooden sailboat docked in a harbor. &  Landscape \\
        \hline
    \end{tabular}
    %}
\end{figure*}

\paragraph{Content Prompts} 
We utilize GPT4 to generate prompts across four meta-categories: human, animal, object, and landscape. Initially, 15/10 prompts (for images/videos) were generated in each category. Recognized as low-quality prompts, semantically repeated prompts, containing style descriptions, and other less informative ones were manually filtered, leading to 5/3 prompts per category. For video prompts, we specifically encouraged the generation of scenarios involving motion. The final prompts in testset are provided in Table \ref{tab:supp_text_prompt_img} and Table \ref{tab:supp_text_prompt_vid}. 

\paragraph{Style References} 
We collect 20 diverse single-reference stylized images and 8 sets of style images with multi-reference(each contains 5 to 7 images in similar styles) from the Internet\footnote{The style references are collected from \url{https://unsplash.com/}, \url{https://unsplash.com/}, \url{https://en.m.wikipedia.org/wiki/}, \url{https://civitai.com/}, \url{https://clipdrop.co/}}. Besides, for the comparison with the Text-to-Image model including Stable Diffusion and the Text-to-Video model including VideoCrafter and Gen-2, we extend them to stylized generation by equipped them with GPT-4v to generate textual style descriptions from style reference. We provide style references and corresponding style descriptions in Table \ref{tab:supp_style_ref} and Figure~\ref{fig:supp_multi_ref}.

\begin{table*}[!h]
    \centering
    \caption{Style references in the testset and corresponding style descriptions generated from GPT-4v\cite{openai2023gpt4v}.}
    \label{tab:supp_style_ref}
    \vspace{-1em}
    \renewcommand{\arraystretch}{1.1}
    %\resizebox{0.99\linewidth}{!}{%
    \small
    \begin{tabular}{>{\centering\arraybackslash}m{2.3cm}m{5.5cm}|>{\centering\arraybackslash}m{2.3cm}m{5.5cm}} 
        %\toprule
        \hline
        Style Reference & Style Descriptions & Style Reference & Style Descriptions \\
        \hline
        \vspace{0.2cm}\includegraphics[width=2.3cm,height=1.7cm,keepaspectratio]{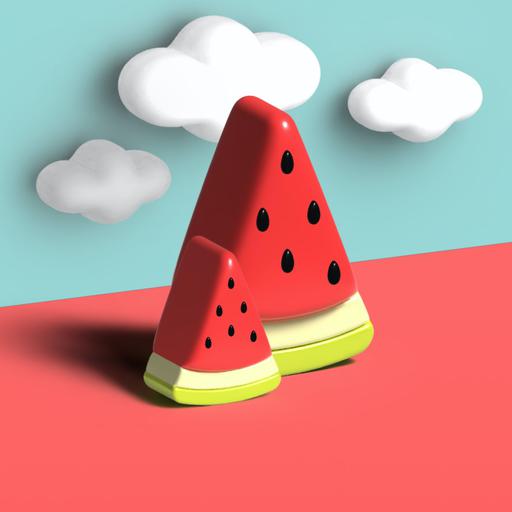}
        & 3D Digital Art, {{\color{blue}{\{prompt\}}}}, whimsical and modern, smooth and polished surfaces, bold and contrasting colors, soft shading and lighting, surreal representation. 
        & \vspace{0.2cm}\includegraphics[width=2.3cm,height=1.7cm,keepaspectratio]{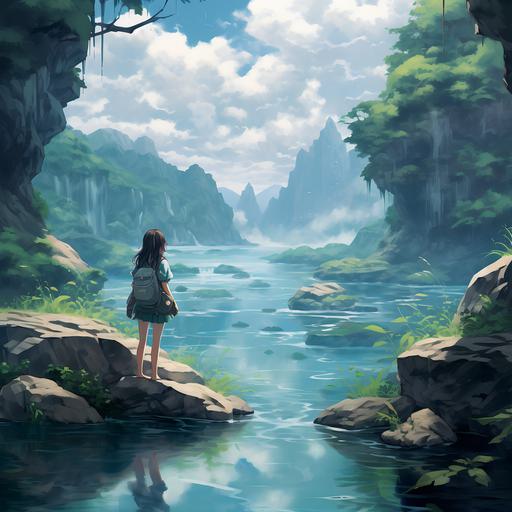}
        & Digital Painting, {\color{blue}{\{prompt\}}}, detailed rendering, vibrant color palette, smooth gradients, realistic light and reflection, immersive natural landscape scene. \\
        \vspace{0.2cm}\includegraphics[width=2.3cm,height=1.7cm,keepaspectratio]{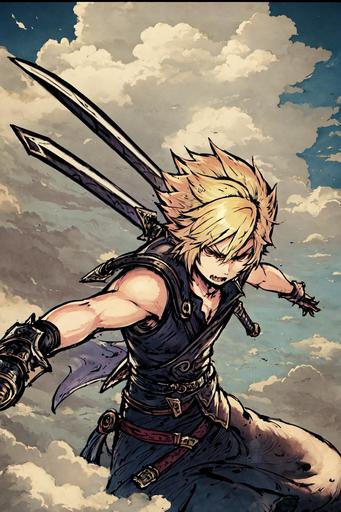}
        & Manga-inspired digital art, {\color{blue}{\{prompt\}}}, dynamic composition, exaggerated proportions, sharp lines, cel-shading, high-contrast colors with a focus on sepia tones and blues. 
        & \vspace{0.2cm}\includegraphics[width=2.3cm,height=1.7cm,keepaspectratio]{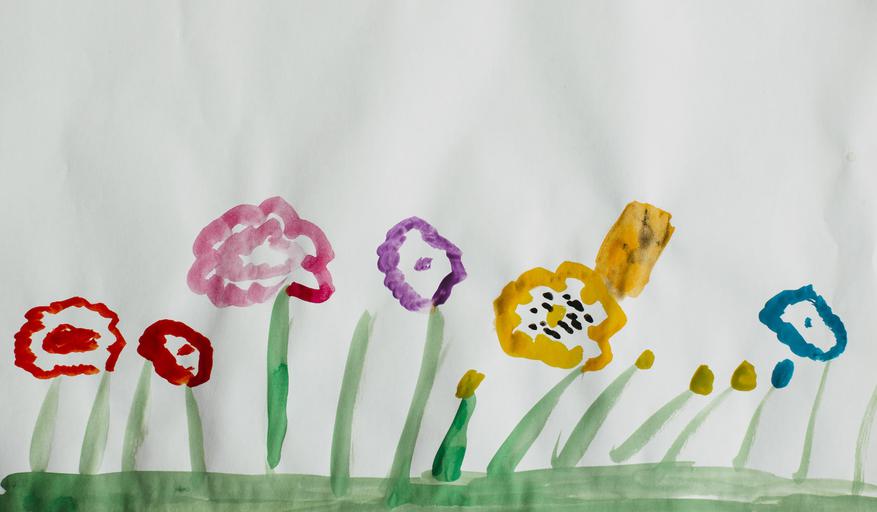}
        & Childlike watercolor, {\color{blue}{\{prompt\}}}, simple brush strokes, primary and secondary colors, bold outlines, flat washes, playful, spontaneous, and expressive. \\
        \vspace{0.2cm}\includegraphics[width=2.3cm,height=1.7cm,keepaspectratio]{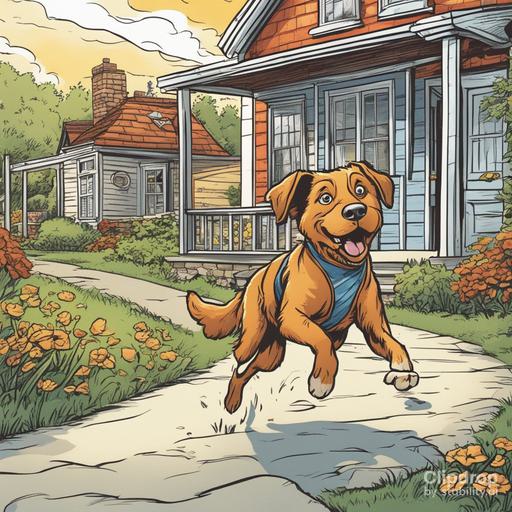}
        & Comic book illustration, {\color{blue}{\{prompt\}}}, digital medium, clean inking, cell shading, saturated colors with a natural palette, and a detailed, textured background. 
        & \vspace{0.2cm}\includegraphics[width=2.3cm,height=1.7cm,keepaspectratio]{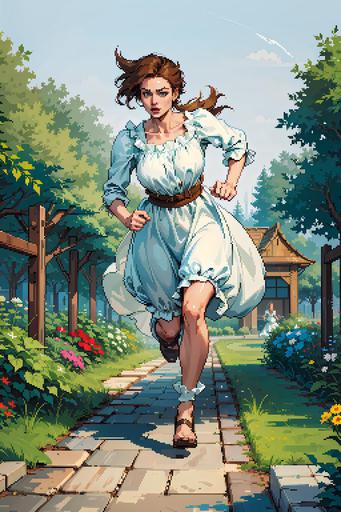}
        & Pixel art illustration, {\color{blue}{\{prompt\}}}, digital medium, detailed sprite work, vibrant color palette, smooth shading, and a nostalgic, retro video game aesthetic. \\
        \vspace{0.2cm}\includegraphics[width=2.3cm,height=1.7cm,keepaspectratio]{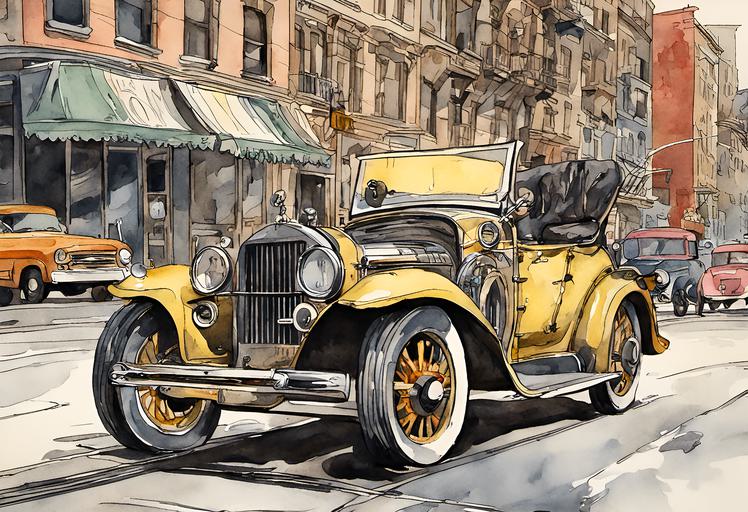}
        & Ink and watercolor on paper, {\color{blue}{\{prompt\}}}, urban sketching style, detailed line work, washed colors, realistic shading, and a vintage feel.
        & \vspace{0.2cm}\includegraphics[width=2.3cm,height=1.7cm,keepaspectratio]{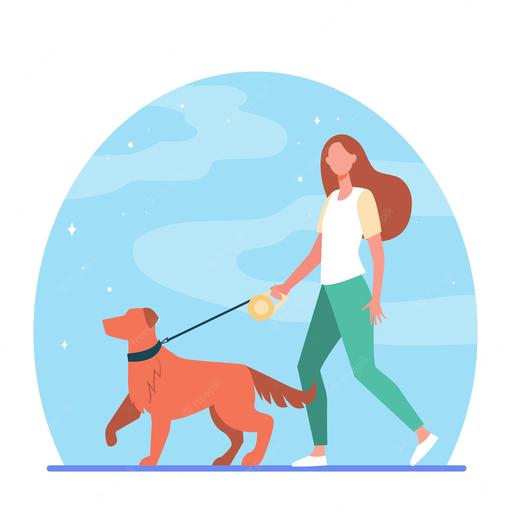}
        & Flat Vector Illustration, {\color{blue}{\{prompt\}}}, simplified shapes, uniform color fills, minimal shading, absence of texture, clean and modern aesthetic. \\
        \vspace{0.2cm}\includegraphics[width=2.3cm,height=1.7cm,keepaspectratio]{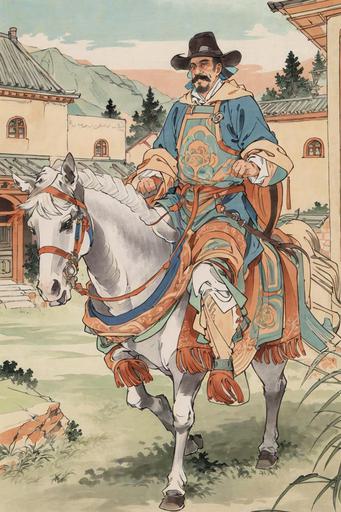}
        & Watercolor and ink illustration, {\color{blue}{\{prompt\}}}, traditional comic style, muted earthy color palette, detailed with a sense of movement, soft shading, and a historic ambiance. 
        & \vspace{0.2cm}\includegraphics[width=2.3cm,height=1.7cm,keepaspectratio]{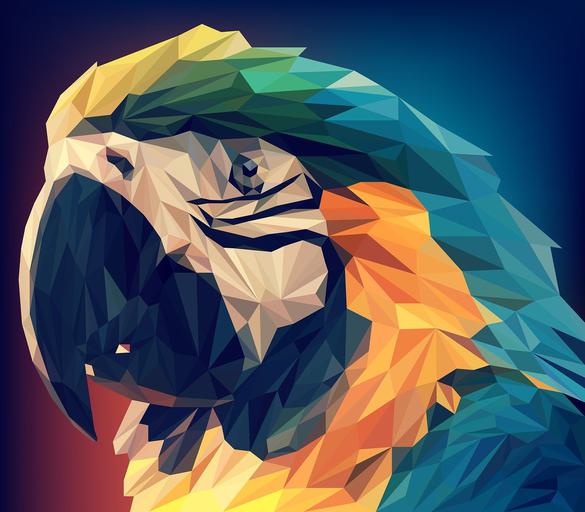}
        & Low Poly Digital Art, {\color{blue}{\{prompt\}}}, geometric shapes, vibrant colors, flat texture, sharp edges, gradient shading, modern graphic style. \\
        \vspace{0.2cm}\includegraphics[width=2.3cm,height=1.7cm,keepaspectratio]{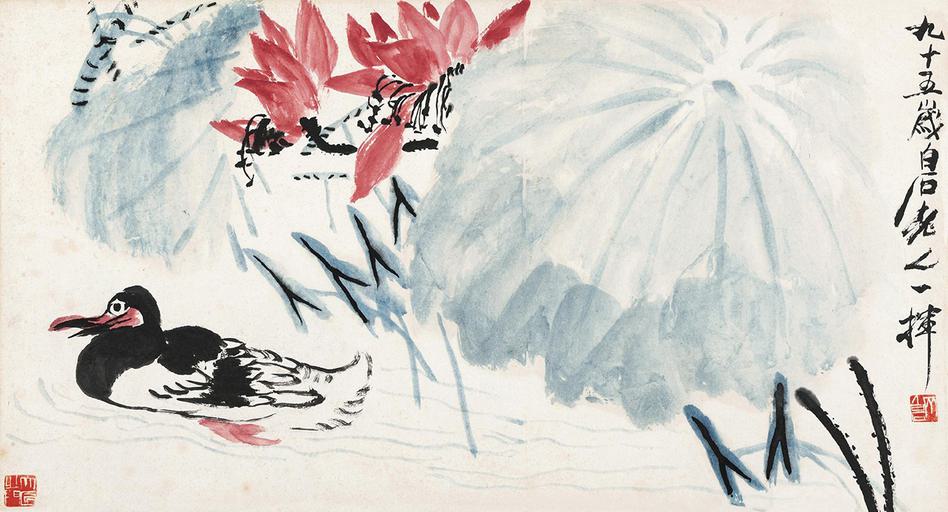}
        & Chinese ink wash painting, {\color{blue}{\{prompt\}}}, minimalistic color use, calligraphic brushwork, emphasis on flow and balance, with poetic inscription.
        & \vspace{0.2cm}\includegraphics[width=2.3cm,height=1.7cm,keepaspectratio]{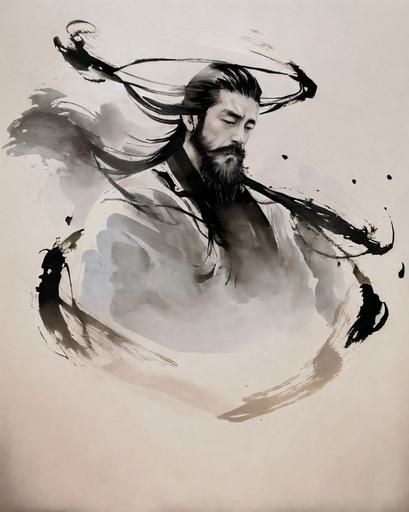}
        & Chinese Ink Wash Painting, {\color{blue}{\{prompt\}}}, monochromatic palette, dynamic brushstrokes, calligraphic lines, with a focus on negative space and movement. \\
        \vspace{0.2cm}\includegraphics[width=2.3cm,height=1.7cm,keepaspectratio]{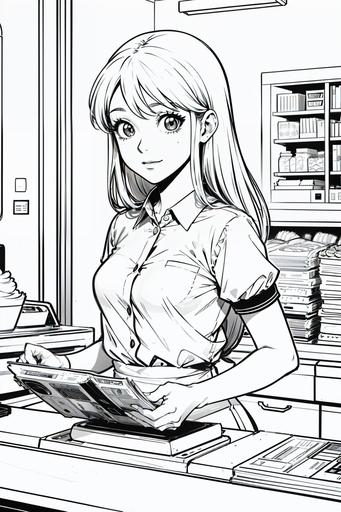}
        & Manga Style, {\color{blue}{\{prompt\}}}, black and white digital inking, high contrast, detailed line work, cross-hatching for shadows, clean, no color.
        & \vspace{0.2cm}\includegraphics[width=2.3cm,height=1.7cm,keepaspectratio]{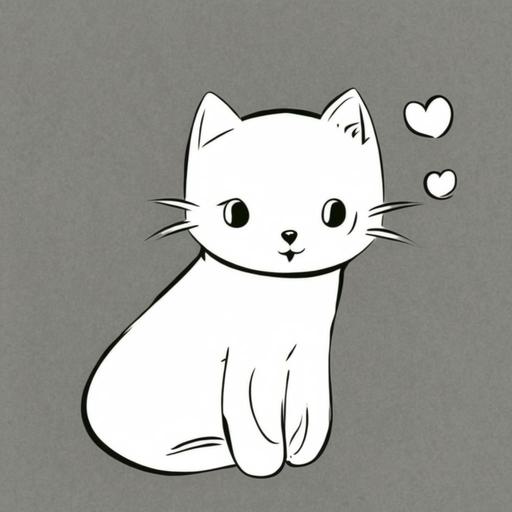}
        & Line Drawing, {\color{blue}{\{prompt\}}}, simple and clean lines, monochrome palette, smooth texture, minimalist and cartoonish representation . \\
        \vspace{0.2cm}\includegraphics[width=2.3cm,height=1.7cm,keepaspectratio]{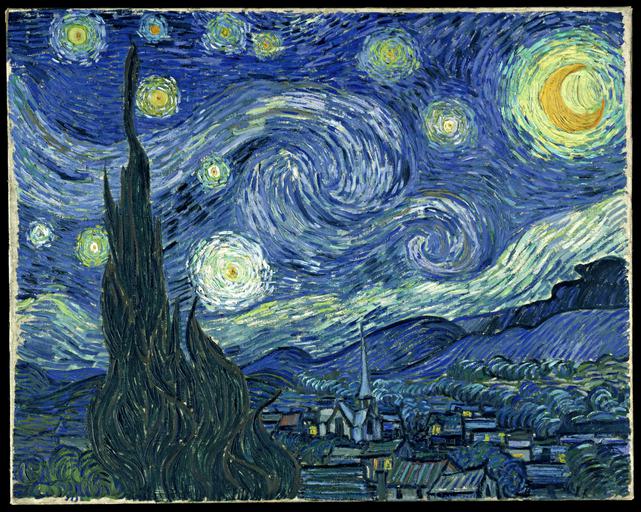}
        & Van Gogh's "Starry Night" style, {\color{blue}{\{prompt\}}}, with expressive, swirling brushstrokes, rich blue and yellow palette, and bold, impasto texture. 
        & \vspace{0.2cm}\includegraphics[width=2.3cm,height=1.7cm,keepaspectratio]{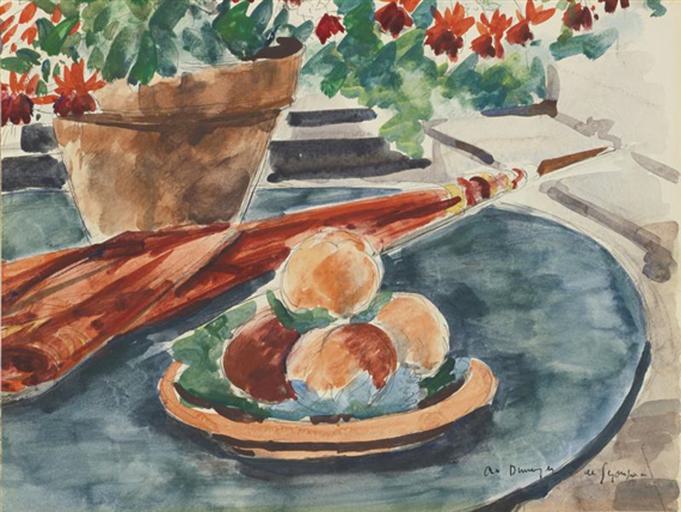}
        & Watercolor Painting, {\color{blue}{\{prompt\}}}, fluid brushstrokes, transparent washes, color blending, visible paper texture, impressionistic style. \\
        \vspace{0.2cm}\includegraphics[width=2.3cm,height=1.7cm,keepaspectratio]{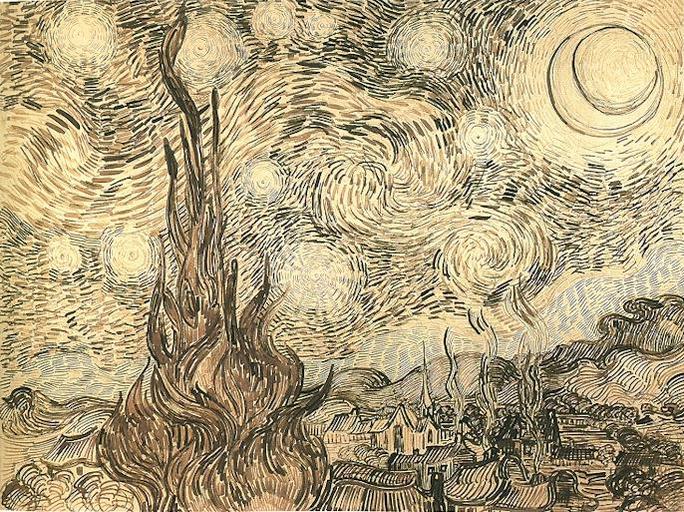}
        & Van Gogh-inspired pen sketch, {\color{blue}{\{prompt\}}}, dynamic and swirling line work, monochromatic sepia tones, textured with a sense of movement and energy. 
        & \vspace{0.2cm}\includegraphics[width=2.3cm,height=1.7cm,keepaspectratio]{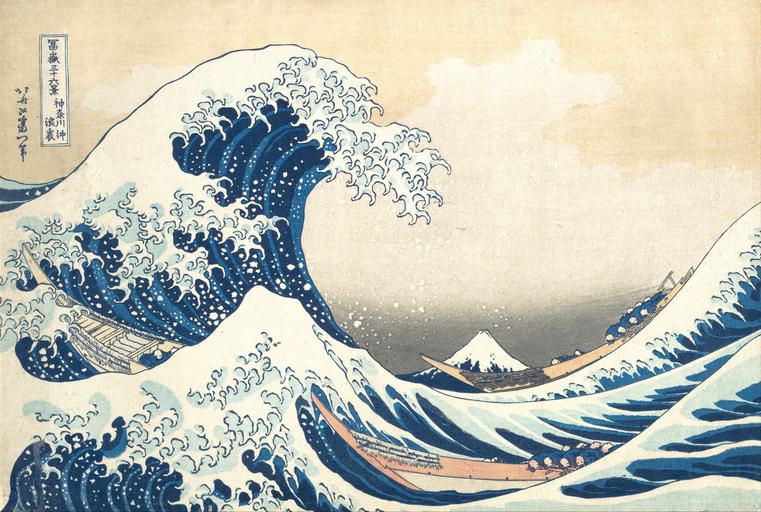}
        & Ukiyo-e Woodblock Print, {\color{blue}{\{prompt\}}}, gradation, limited color palette, flat areas of color, expressive line work, stylized wave forms, traditional Japanese art. \\
        \vspace{0.2cm}\includegraphics[width=2.3cm,height=1.7cm,keepaspectratio]{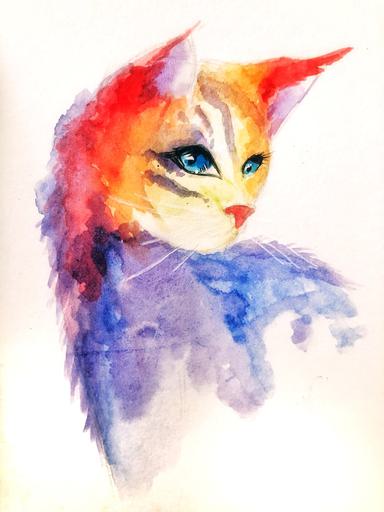}
        & Watercolor Painting, {\color{blue}{\{prompt\}}}, fluid washes of color, wet-on-wet technique, vibrant hues, soft texture, impressionistic portrayal.
        & \vspace{0.2cm}\includegraphics[width=2.3cm,height=1.7cm,keepaspectratio]{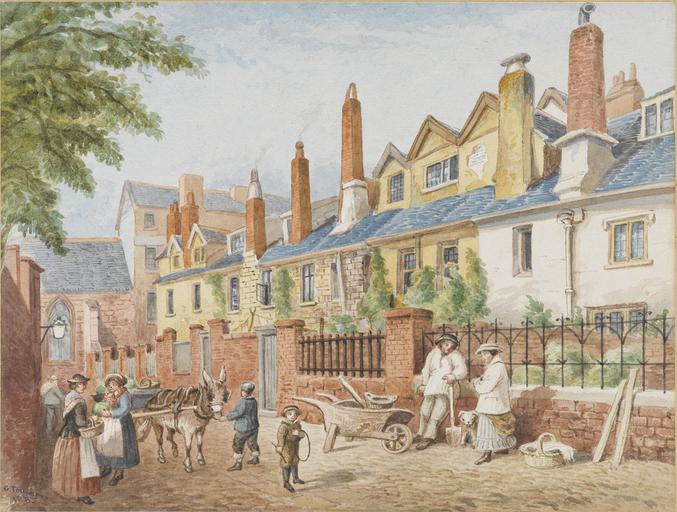}
        & Victorian watercolor, {\color{blue}{\{prompt\}}}, fine detail, soft pastel hues, gentle lighting, clear texture, with a quaint, realistic portrayal of everyday life. \\
        \hline
    \end{tabular}
    %}
\end{table*}

\begin{figure*}[p]
    \centering
    \includegraphics[width=0.72\linewidth]{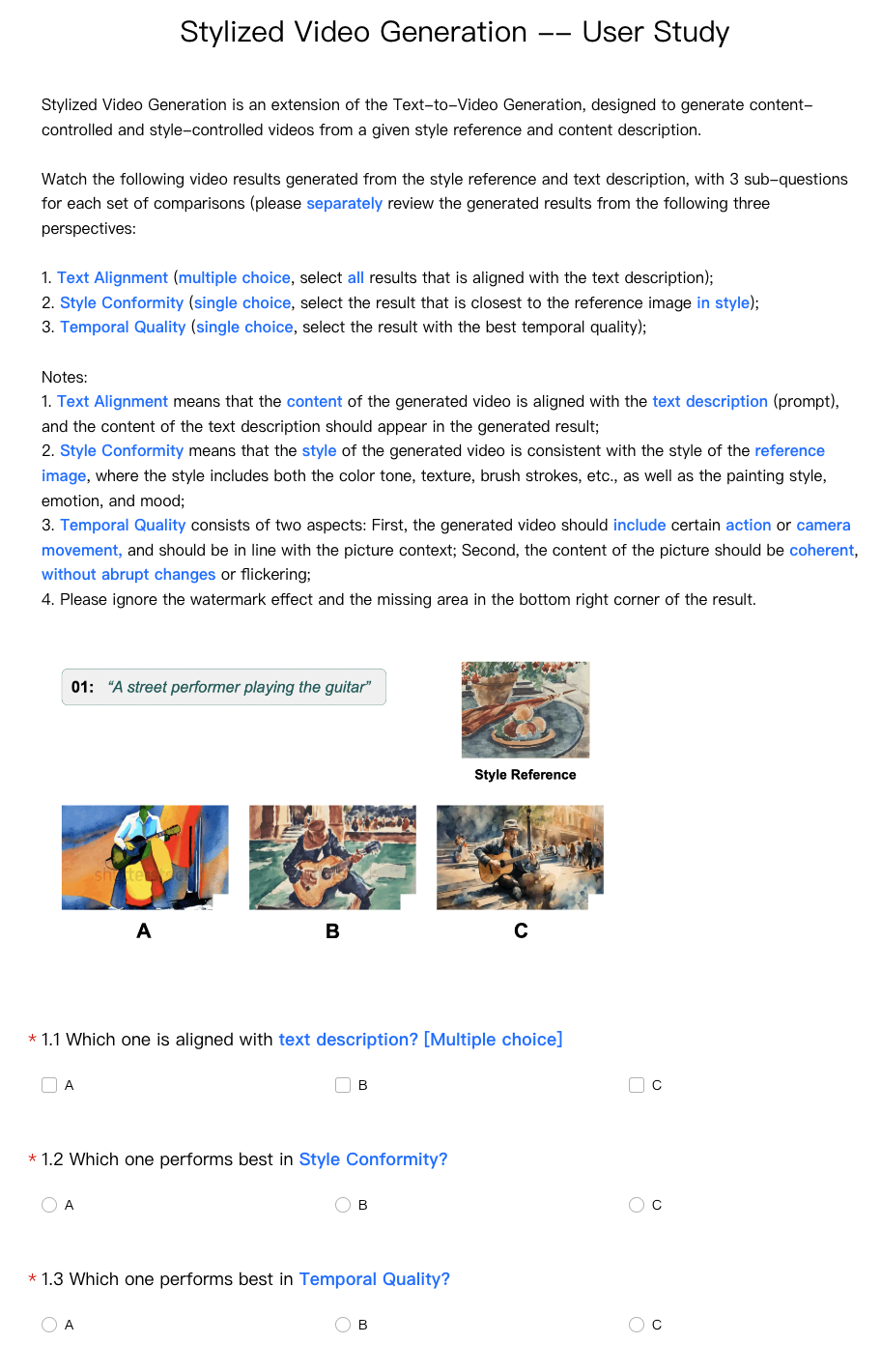}
    \caption{User Preference Study Interface} 
    \label{fig:supp_user_study}
\end{figure*}

%
% For text alignment, participants were instructed to select all results that aligned with the text description. For style conformity, they were asked to choose the result that most closely resembled the reference image in style. For temporal quality, participants were instructed to select the result with the best temporal quality, which included both obvious actions and temporal consistency. 
%

\section{Evaluation Details}
\label{sec:supp_eval}
\subsection{Evaluation Metrics}

We employ CLIP-based similarity scores to evaluate text alignment and style conformity, as is commonly done by existing methods. Additionally, we include two metrics to measure the temporal consistency of generated videos, i.e., \textit{CLIP-Temp} and \textit{Warping Error}. A detailed calculation process for each metric is presented below.

\paragraph{CLIP-Text} We utilize the pretrained CLIP-ViT-H-14-laion2B-s32B-b79K\footnote{\url{https://huggingface.co/laion/CLIP-ViT-H-14-laion2B-s32B-b79K}} as a feature extractor(also for \textit{CLIP-Style} and \textit{CLIP-Temp}), which is trained on LAION-2B and demonstrates enhanced performance across various datasets. We extract the frame-wise image embeddings from the generated results and text embeddings from the input content prompts, then compute their average cosine similarity. The overall \textit{CLIP-Text} is calculated as:

\begin{equation}
    S_{text} = \frac{1}{M}\sum_{i = 1}^{M}(\frac{1}{T}\sum_{t=1}^{T}\frac{emb(x_t^i) \cdot emb(p^i)}{\lVert emb(x_t^i) \rVert \cdot \lVert emb(p^i) \rVert})
\end{equation}
where $M$ represents the total number of testing videos and $T$ represents the total number of frames in each video($T = 1$ for image generation), $emb(x_t^i)$ and $emb(p^i)$ indicate the CLIP embedding of the $t$-th frame of the $i$-th video $x_t^i$ and the corresponding prompt $p^i$, respectively.

\paragraph{CLIP-Style} Similarly, we extract the frame-wise image embeddings $emb(x_t^i)$ from the generated results and image embeddings $s^i$ from the input style reference. The overall \textit{CLIP-Text} is calculated as:

\begin{equation}
    S_{style} = \frac{1}{M}\sum_{i = 1}^{M}(\frac{1}{T}\sum_{t=1}^{T}\frac{emb(x_t^i) \cdot emb(s^i)}{\lVert emb(x_t^i) \rVert \cdot \lVert emb(s^i) \rVert})
\end{equation}

\paragraph{CLIP-Temp} Considering the semantic consistency between every two frames, we extract the frame-wise CLIP image embeddings and compute the cosine similarity between each of the two frames, as follows:

\begin{equation}
    S_{temp} = \frac{1}{M}\sum_{i = 1}^{M}(\frac{1}{T-1}\sum_{t=1}^{T - 1}\frac{emb(x_t^i) \cdot emb(x_{t + 1}^i)}{\lVert emb(x_t^i) \rVert \cdot \lVert emb(x_{t + 1}^i) \rVert})
\end{equation}

\paragraph{Warping Error} For the warping error, we first obtain the optical flow between each two frames using RAFT-small\cite{teed2020raft}, a pre-trained optical flow estimation network. Subsequently, we compute the pixel-wise differences between the warped image and the predicted image, as follows:

\begin{equation}
    W_{error} = \frac{1}{M}\sum_{i = 1}^{M}(\frac{1}{T-1}\sum_{t=1}^{T - 1} \lVert x_t^i - warp(x_{t + 1}^i, flow(x_{t + 1}^i, x_{t}^i))\rVert)
\end{equation}
where $warp(x_{t + 1}^i, flow(x_{t + 1}^i, x_{t}^i))$ represents the warped frame of $x_{t + 1}^i$ using the optical flow between frame $x_{t + 1}^i$ and frame $x_{t}^i$.

\subsection{User Study}

In this subsection, we provide a detailed introduction about our user study. We randomly selected 15 single-reference style-text pairs to compare the generated results among VideoCrafter~\cite{chen2023videocrafter}, Gen-2~\cite{Gen-2}, and our proposed method. Given that videocomposer~\cite{wang2024videocomposer} directly replicates the style reference and is minimally influenced by the prompt in most cases, we excluded it from the comparison in the user study. Additionally, we randomly chose 10 multi-reference style-text pairs for the comparison between AnimateDiff~\cite{guo2023animatediff} (multiple style-specific models) and our method (a generic model). To ensure a blind comparison, we randomized the order of options for each question and masked the possible model watermark in the lower right corner.

The designed user preference interface is illustrated in Figure~\ref{fig:supp_user_study}. We invited 15 users of normal eyesight to evaluate the generated results in three aspects: text alignment, style conformity, and temporal quality.  The instructions and questions are provided as below.
%
% For text alignment, participants were instructed to select all results that aligned with the text description. For style conformity, they were asked to choose the result that most closely resembled the reference image in style. For temporal quality, participants were instructed to select the result with the best temporal quality, which included both obvious actions and temporal consistency. 
%
Consequently, a total of 1125 votes are collected.\vspace{0.6em}

\noindent\textbf{Instructions.}

\begin{itemize}[leftmargin=2.0em]
    \item \textbf{Task}: Watch the following video results generated from the style reference and text description, with 3 sub-questions for each set of comparisons (please \textbf{separately} review the generated results from the following three perspectives:

    \begin{itemize}[leftmargin=2.0em]
        \item \textbf{Text Alignment} (multiple choice, means that the content of the generated video is aligned with the text description(prompt), and the content of the text description should appear in the generated result);
        \item \textbf{Style Conformity} (single choice, means that the style of the generated video is consistent with the style of the reference image, where the style includes both the color tone, texture, brush strokes, etc., as well as the painting style, emotion, and mood);
        \item \textbf{Temporal Quality} (single choice, consists of two aspects: First, the generated video should include certain action or camera movement, and should be in line with the picture context; Second, the content of the picture should be coherent, without abrupt changes or flickering);
    \end{itemize}
    
    \item Please ignore the watermark effect and the missing area in the bottom right corner of the result.
    % \begin{itemize}
    %     \item Text Alignment means that the content of the generated video is aligned with the text description(prompt), and the content of the text description should appear in the generated result;
    %     \item Style Conformity means that the style of the generated video is consistent with the style of the reference image, where the style includes both the color tone, texture, brush strokes, etc., as well as the painting style, emotion, and mood;
    %     \item Temporal Quality consists of two aspects: First, the generated video should include certain action or camera movement, and should be in line with the picture context; Second, the content of the picture should be coherent, without abrupt changes or flickering;
    %     \item Please ignore the watermark effect and the missing area in the bottom right corner of the result.
    % \end{itemize}
\end{itemize}

% \vspace{0.6em}
\noindent\textbf{Questions.}

\begin{itemize}[leftmargin=2.0em]
    \item Which one is aligned with text description? \textbf{[Multiple choice]}
    \item Which performs best in Style Conformity? \textbf{[Single choice]}
    \item Which performs best in Temporal Quality? \textbf{[Single choice]}
\end{itemize}

%%%%%%%%%%%%%%%%%%%%%%%%%%%%%%%%%
\section{Extended Comparison}
\label{sec:supp_extend_comp}

\subsection{Multi-reference Stylized Image Generation}

\begin{figure*}[!h]
    \centering
    \includegraphics[width=0.9\linewidth]{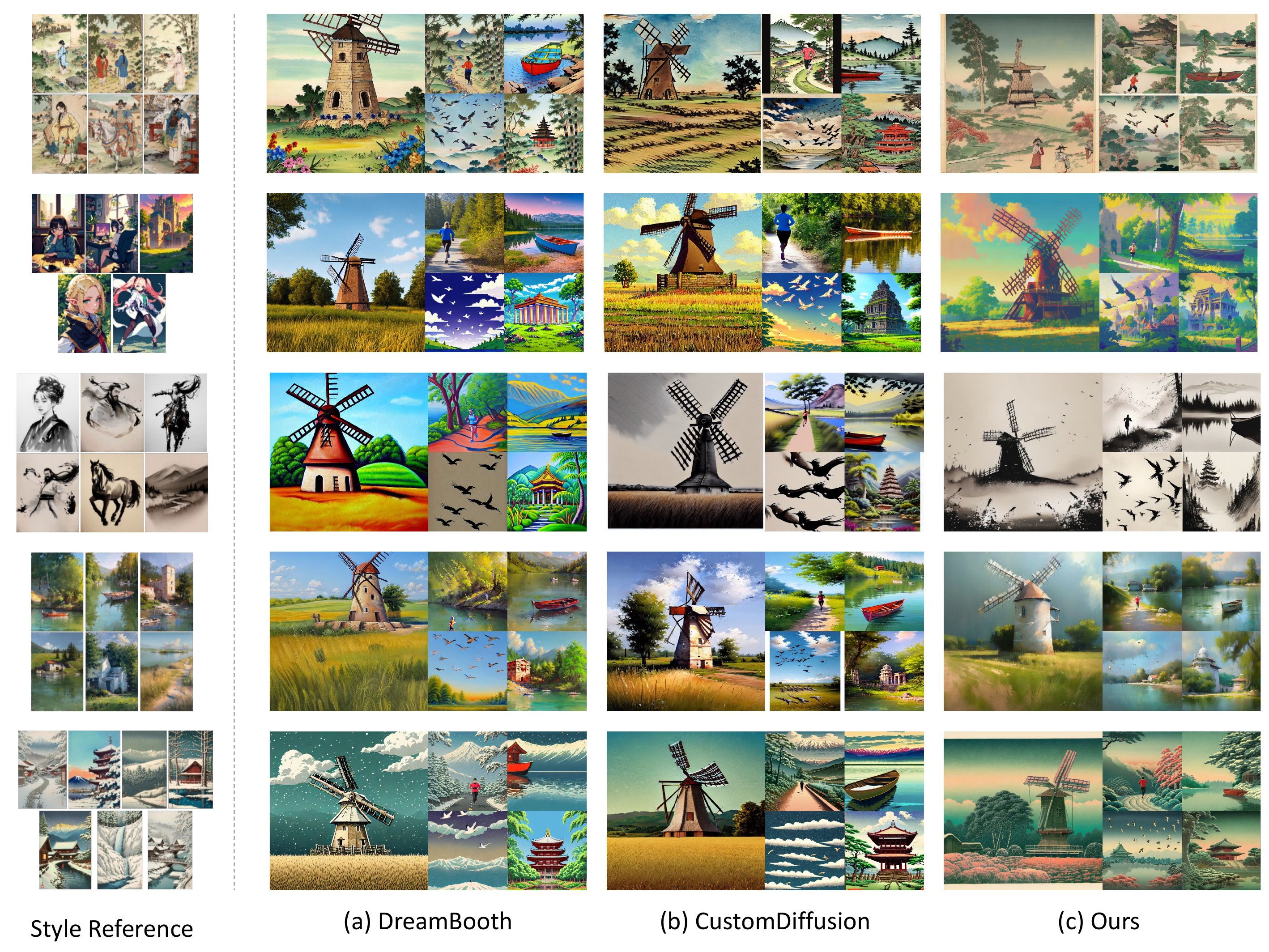}
    \vspace{-1.5em}
    \caption{Visual comparison on mulit-reference stylized T2I generation. Testing prompts: (i) \textit{A rustic windmill in a field.}; (ii) \textit{A person jogging along a scenic trail.}; (iii) \textit{A flock of birds flying gracefully in the sky.}; (iv) \textit{A rowboat docked on a peaceful lake.}; (v) \textit{An ancient temple surrounded by lush vegetation.}} 
    \label{fig:supp_comp_multi_ref}
\end{figure*}

We conduct comparisons of multi-reference stylized image generation with Dreambooth~\cite{dreambooth} and CustomDiffusion~\cite{customdiffusion}, both of which support generating images in specific styles by finetuning on the reference images. Figure~\ref{fig:supp_multi_ref} and Table~\ref{tab:supp_multi_ref} present the visual and quantitative results respectively, demonstrating that our method surpasses all competitors in terms of style conformity for multi-reference stylized generation. Although Dreambooth and CustomDiffusion exhibit competitive performance in certain cases, their stylized generation abilities tend to vary with different prompts, i.e., struggling to maintain consistent visual styles across arbitrary prompts. It is possibly because several images are insufficient to allow the model to disentangle the contents and styles, thus harming the generalization performance.
Besides, the requirement for finetuning during the testing process also undermines their flexibility. In contrast, our method efficiently generates high-quality stylized images that align with the prompts and conform the style of reference images without additional finetuning costs.

\begin{table}[!h]
\centering
\caption{Quantitative comparison on Multi-reference style-guided T2I generation. \textbf{Bold}: Best.}
\label{tab:supp_multi_ref}
\vspace{-1.0em}
%\setlength{\tabcolsep}{3.0pt}
% \renewcommand\arraystretch{1.5}
%\resizebox{\linewidth}{!}{
  \begin{tabular}{ccccc} % {@{}lc@{}}
    % \toprule
    %  \hline
    %  & \multicolumn{5}{|c}{\textbf{CLIP scores}} \\
    \hline
    Methods & Dreambooth & CustomDiffsion  & Ours \\
    \hline
    \texttt{Text} $\uparrow$ & 0.2868 & \textbf{0.2986} & 0.2924 \\
    \texttt{Style} $\uparrow$ & 0.4270 & 0.4441 & \textbf{0.5333} \\
    \hline
  \end{tabular}
% }
\end{table}

% \subsection{Comparison with AnimateDiff}
% \TODO{(11.23) 1. We have pretrained a good personalized T2I model, but it has degraded in performance of style and temporal. 2. Our performance is comparable with AnimateDiff + their good personalized model.}

\vspace{-1.0em}
\subsection{Comparison with StyleDrop}

\begin{figure*}[!t]
    \centering
    \includegraphics[width=\linewidth]{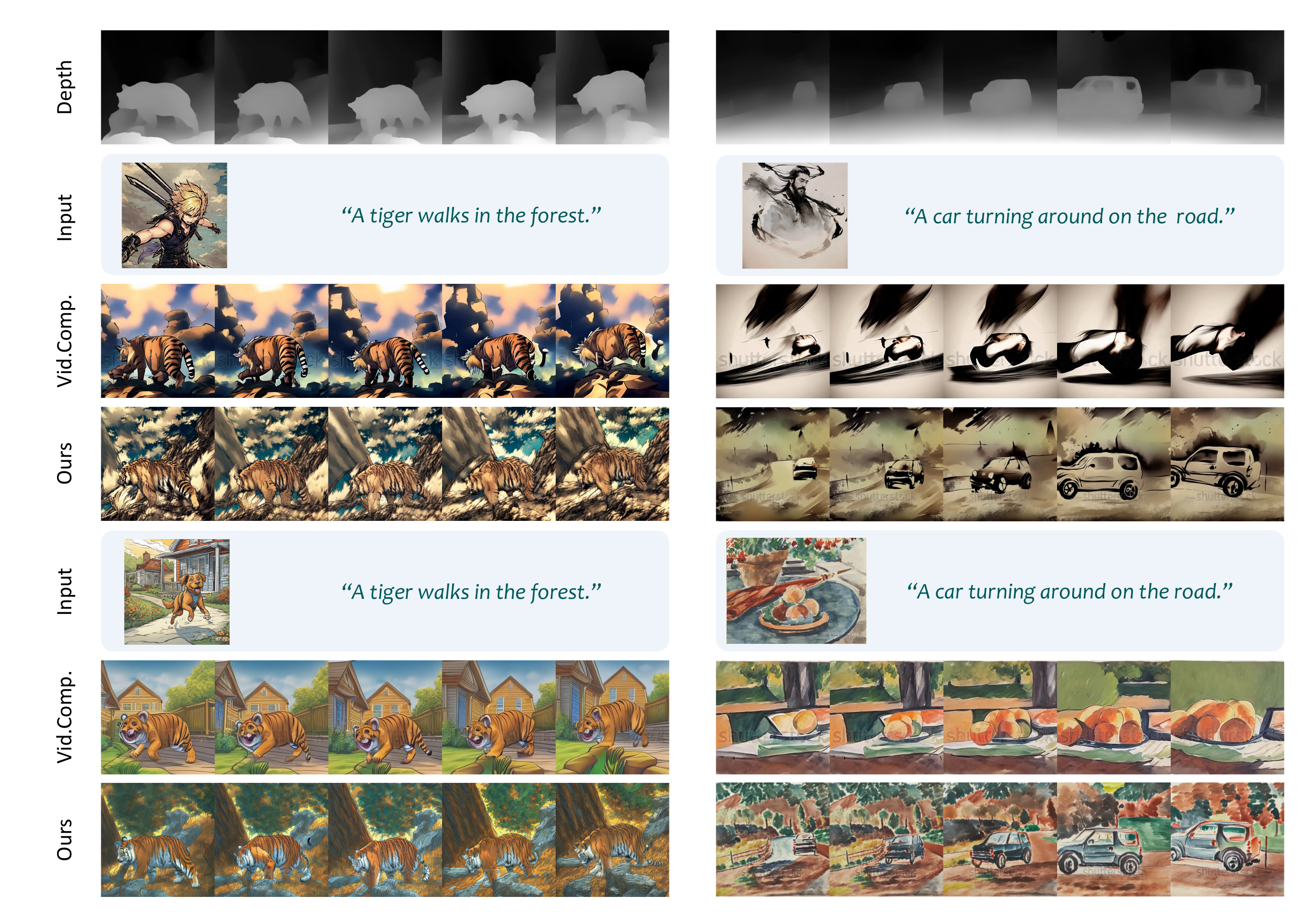}
    \vspace{-2em}
    \caption{Visual comparison on stylized video generation with additional depth guidance. Vid.Comp.: VideoComposer} 
    \label{fig:supp_depth}
\end{figure*}

\begin{figure}[!h]
    \centering
    \vspace{-0.8em}
    \includegraphics[width=0.8\linewidth]{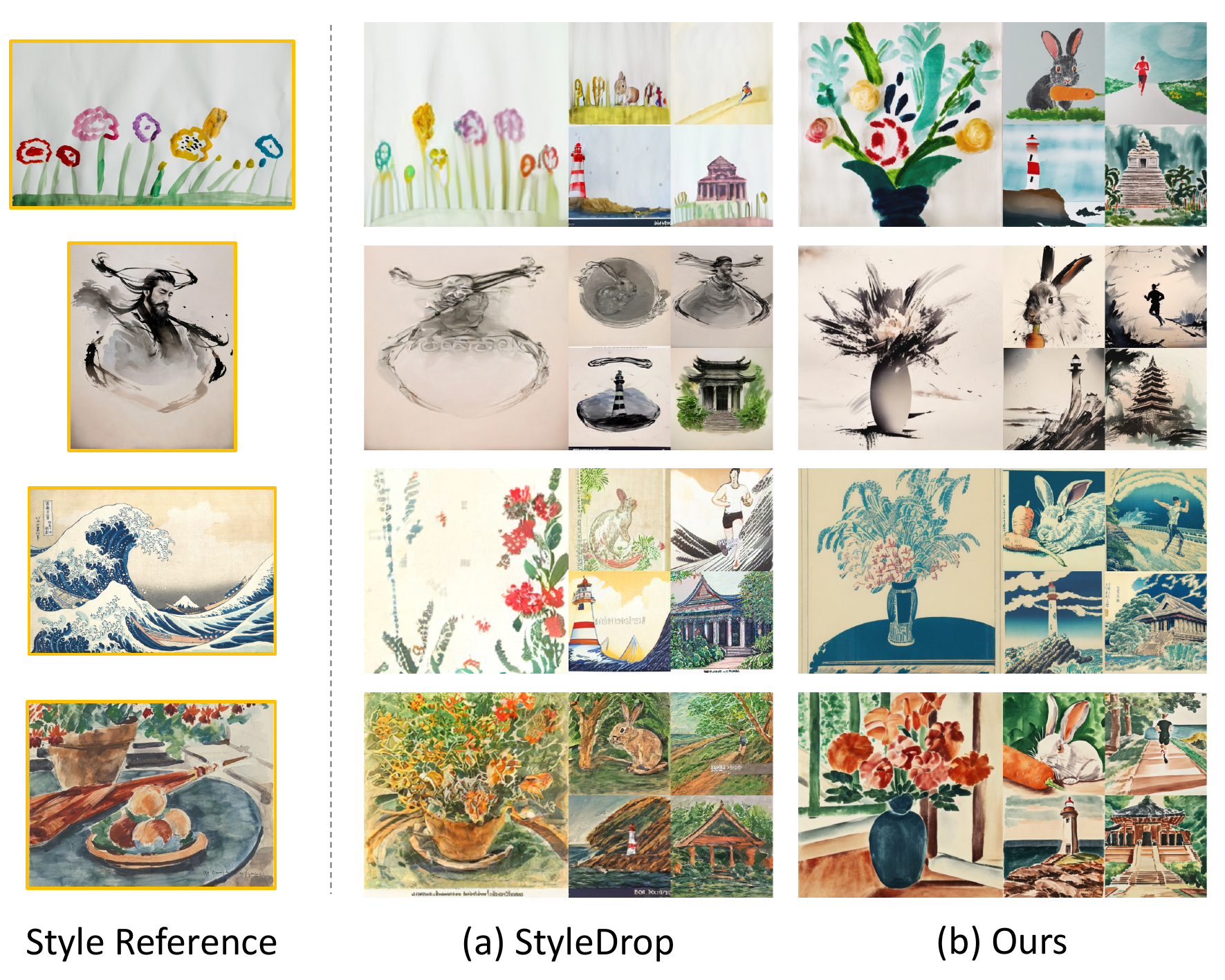}
    \vspace{-1.2em}
    \caption{Visual comparison between StyleDrop and our proposed method. Testing prompts: (i) \textit{A woman reading a book in a park.}; (ii) \textit{A person jogging along a scenic trail.}; (iii) \textit{A colorful butterfly resting on a flower.}; (iv) \textit{A rabbit nibbling on a carrot.}; (v) \textit{A telescope pointed at the stars.}} 
    \label{fig:supp_comp_styledrop}
\end{figure}

Here we present a supplementary comparison with StyleDrop\cite{sohn2023styledrop}. StyleDrop proposes a versatile method for synthesizing images that faithfully follow a specific style using a text-to-image model. Owing to the absence of an official StyleDrop implementation, we have excluded the comparison with StyleDrop from the main text. Instead, we include a comparison with an unofficial StyleDrop implementation\footnote{\url{https://github.com/aim-uofa/StyleDrop-PyTorch}} here. We train StyleDrop based on StableDiffusion 2.1 for 1000 steps with a batch size of 8 and a learning rate of $3 \times 10^{-4}$. The quantitative and qualitative results are presented in Table~\ref{tab:supp_comp_styledrop} and Figure~\ref{fig:supp_comp_styledrop} respectively. Results show that compared to StyleDrop, our proposed method more effectively captures the visual characteristics of a user-provided style and combines them with various prompts in a flexible manner.

\begin{table}[!h]
\centering
%\setlength{\tabcolsep}{3.0pt}
% \renewcommand\arraystretch{1.5}
% \vspace{-1em}
\caption{Quantitative comparison with StyleDrop.}
\vspace{-1em}
\label{tab:supp_comp_styledrop}
\resizebox{0.6\linewidth}{!}{
  \begin{tabular}{ccc} % {@{}lc@{}}
    % \toprule
    %  \hline
    %  & \multicolumn{5}{|c}{\textbf{CLIP scores}} \\
    \hline
    Methods & StyleDrop & Ours(SD 2.1) \\
    \hline
    \texttt{Text} $\uparrow$ & 0.2389 & 0.3028 \\
    \texttt{Style} $\uparrow$ & 0.3962 & 0.4836 \\
    \hline
  \end{tabular}
}
\end{table}

\subsection{Comparison with Style Transfer Methods}

In this section, we perform comparisons with a particular set of competitors, i.e., style transfer methods. As style transfer methods require extra source images or videos for content provision, we utilize text-to-image models(Stable Diffusion 2.1) and text-to-video models(VideoCrafter) to initially generate visual content. Subsequently, we apply existing style transfer methods based on the reference to produce the stylized output. For stylized image generation, we opt for T2I + CAST\cite{zhang2022domain}, while for stylized video generation, we choose T2V + MCCNet\cite{deng2021arbitrary}. The results of this comparison are illustrated in the Table~\ref{tab:supp_comp_cast} and Table~\ref{tab:supp_comp_mccnet}. 
Both two competitors underperform our approach.
The reasons are in two aspects: (i) Style transfer methods mainly work well on transferring tones and local textures while fall short in transferring semantic style features; (ii) Style transfer method cannot change the structure of the content image, which hinders the generation of styles that associated with special geometry features, e.g. layouts of logo style and 3D render style.

\begin{table}[!h]
\centering
%\setlength{\tabcolsep}{3.0pt}
% \renewcommand\arraystretch{1.5}
% \vspace{-1em}
\caption{Quantitative comparison with image style transfer method.}
\vspace{-1em}
\label{tab:supp_comp_cast}
%\resizebox{0.6\linewidth}{!}{
  \begin{tabular}{ccc} % {@{}lc@{}}
    % \toprule
    %  \hline
    %  & \multicolumn{5}{|c}{\textbf{CLIP scores}} \\
    \hline
    Methods & T2I + CAST & Ours(SD 2.1) \\
    \hline
    \texttt{Text} $\uparrow$ & 0.3027 & 0.3028 \\
    \texttt{CLIP-Style} $\uparrow$ & 0.3549 & 0.4836 \\
    % \texttt{DINO-Style} $\uparrow$ & 0.2645 & 0.3652 \\
    \hline
  \end{tabular}
% }
\end{table}

\begin{table}[!h]
\centering
%\setlength{\tabcolsep}{3.0pt}
% \renewcommand\arraystretch{1.5}
% \vspace{-1em}
\caption{Quantitative comparison with video style transfer method.}
\vspace{-1em}
\label{tab:supp_comp_mccnet}
% \resizebox{0.6\linewidth}{!}{
  \begin{tabular}{ccc} % {@{}lc@{}}
    % \toprule
    %  \hline
    %  & \multicolumn{5}{|c}{\textbf{CLIP scores}} \\
    \hline
    Methods & T2I + MCCNet & Ours(VideoCrafter) \\
    \hline
    \texttt{Text} $\uparrow$ & 0.2487 & 0.2726 \\
    \texttt{Style} $\uparrow$ & 0.2858 & 0.4531 \\
    \texttt{Temporal} $\uparrow$ & 0.9577 & 0.9892 \\
    \hline
  \end{tabular}
% }
\end{table}

%%%%%%%%%%%%%%%%%%%%%%%%%%%%%%%%%
\section{Extended Ablation Study}
\label{sec:supp_extend_ablation}
% \addcontentsline{toc}{chapter}{D. Extended Ablation Study}

\begin{figure}[h]
    \centering
    \includegraphics[width=\linewidth]{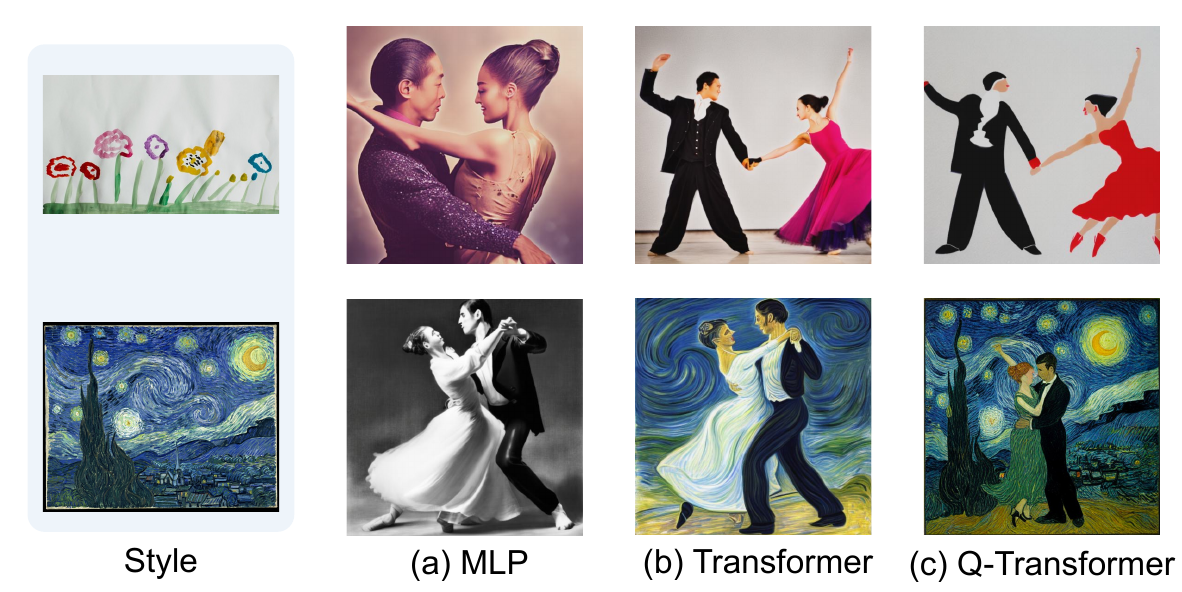}
    \vspace{-1em}
    \caption{Visual comparison between different adapter architectures. Prompt: \textit{A couple dancing gracefully together.}}
    \label{fig:supp_ablation_arch}
\end{figure}

Since we have made ablation studies on some key design in the main text(i.e., Dual Cross-Attention, Data Augmentation, and Adaptive Style-Content Fusion), we provide additional comparison between different style adapter architechures, including MLP, Transformer, and Query Transformer(ours). Quantitative results and visual comparisons in Figure \ref{fig:supp_ablation_arch} and Table \ref{tab:supp_ablation_arch} show that Query Transformer excels over the other alternatives.

\begin{table}[!h]
% \vspace{-1em}
\centering
\caption{Ablation studies on style feature extractor architecture. The performance is evaluated on the style-guided T2I generation.}
\label{tab:supp_ablation_arch}
\vspace{-1em}
  \begin{tabular}{cccc} % {@{}lc@{}}
    % \toprule
    \hline
    Alternatives & MLP & Transformer & Q-Former (Ours) \\
    \hline
    Text $\uparrow$ & 0.3415 & 0.3221 & 0.3028 \\
    Style $\uparrow$ & 0.2843 & 0.4149 & 0.4836 \\
    \hline
  \end{tabular}
  
\end{table}

%%%%%%%%%%%%%%%%%%%%%%%%%%%%%%%%%
\section{Application Extension}
\label{sec:supp_app}

In this section, we further explore the compatibility with additional controllable conditions, e.t., depth. Following the approach of structure control in Animate-A-Story\cite{he2023animate}, we introduce video structure control by integrating a well-trained depth adapter into the base T2V model. Note that StyleCrafter and depth-adapter are trained independently, the only operation we take is to combine the both during the inference stage. Instead of employing DDIM Inversion to ensure consistency, we generate the videos from random noise. The visual comparison with VideoComposer\cite{wang2024videocomposer} is present in Figure~\ref{fig:supp_depth}. VideoComposer struggles to produce results faithful to text descriptions when faced with artistic styles, such as the "boat" mentioned in the prompt. In contrast, our method not only supports collaboration with depth guidance, but also generates videos with controllable content, style, and structure.

% %%%%%%%%%%%%%%%%%%%%%%%%%%%%%%%%%
% \section{Further Exploration}
% \TODO{(11.23-11.24) We have explored an enhanced version trained with a few stylized videos.}

%%%%%%%%%%%%%%%%%%%%%%%%%%%%%%%%%
\section{More Results}
\label{sec:supp_more_result}

\begin{figure*}[p]
    \vspace*{\fill}
    \centering
    \includegraphics[width=\linewidth]{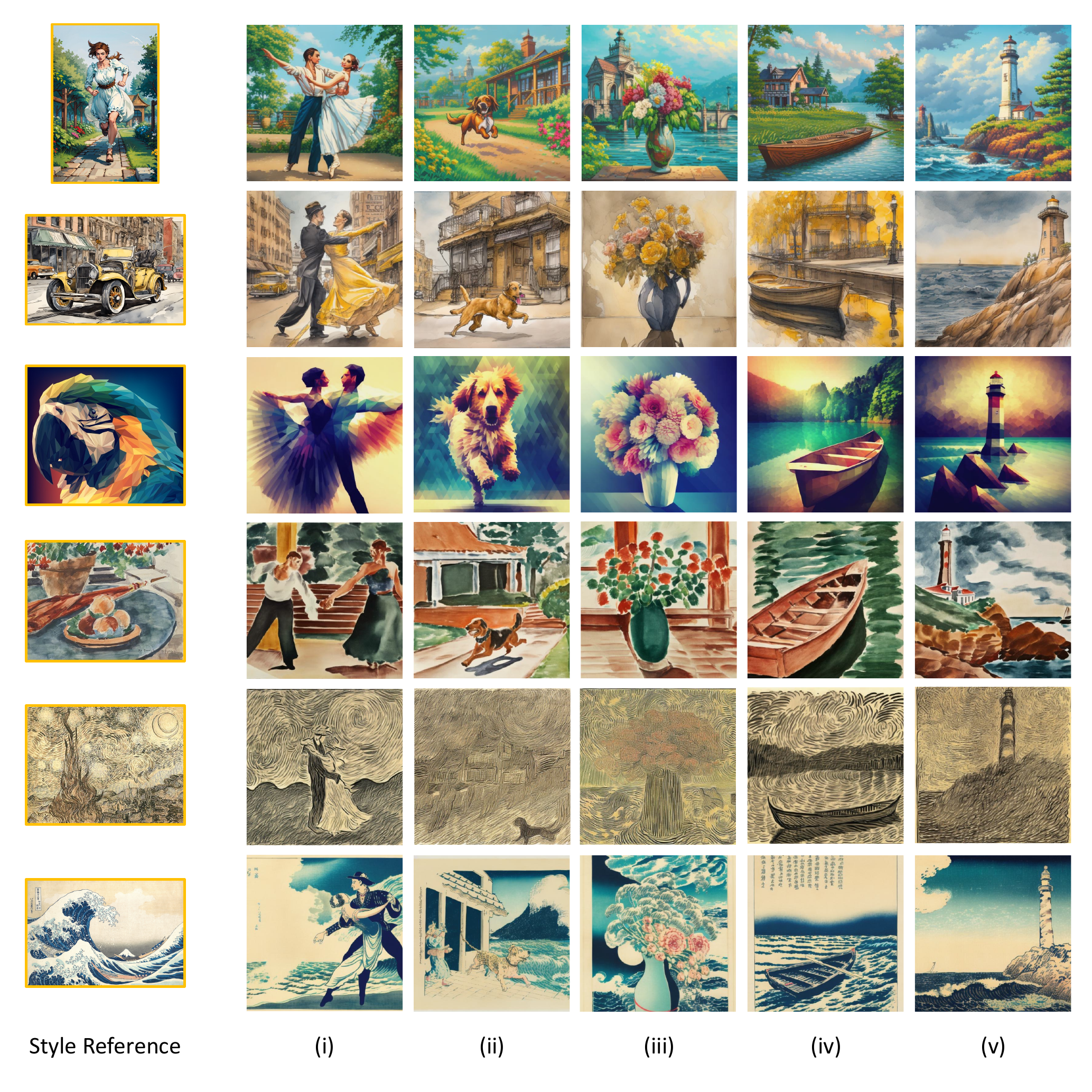}
    \caption{More Results of \textbf{StyleCrafter(SD2)} on Style-Guided Text-to-Image Generation. Prompts: (i) A couple dancing gracefully together. (ii) A dog is running in front of a house. (iii) A bouquet of flowers in a vase. (iv) A rowboat docked on a peaceful lake. (v) A lighthouse standing tall on a rocky coast.} 
    \label{fig:supp_more_result_img_sd}
    \vspace*{\fill}
\end{figure*}

\begin{figure*}[p]
    \vspace*{\fill}
    \centering
    \includegraphics[width=\linewidth]{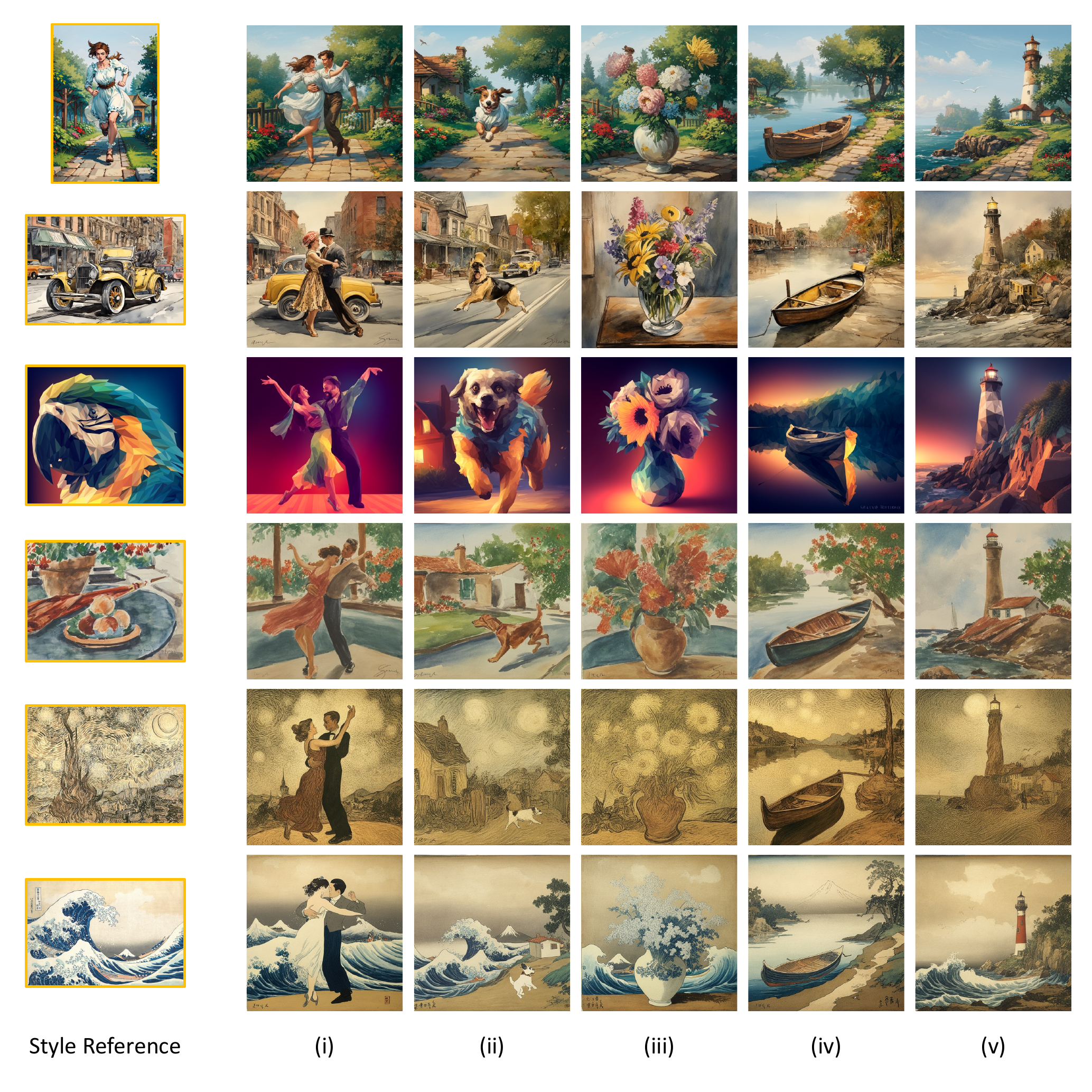}
    \caption{More Results of \textbf{StyleCrafter(SDXL)} on Style-Guided Text-to-Image Generation. Prompts: (i) A couple dancing gracefully together. (ii) A dog is running in front of a house. (iii) A bouquet of flowers in a vase. (iv) A rowboat docked on a peaceful lake. (v) A lighthouse standing tall on a rocky coast.} 
    \label{fig:supp_more_result_img_sdxl}
    \vspace*{\fill}
\end{figure*}

\begin{figure*}[t]
    \centering
    \includegraphics[width=0.9\linewidth]{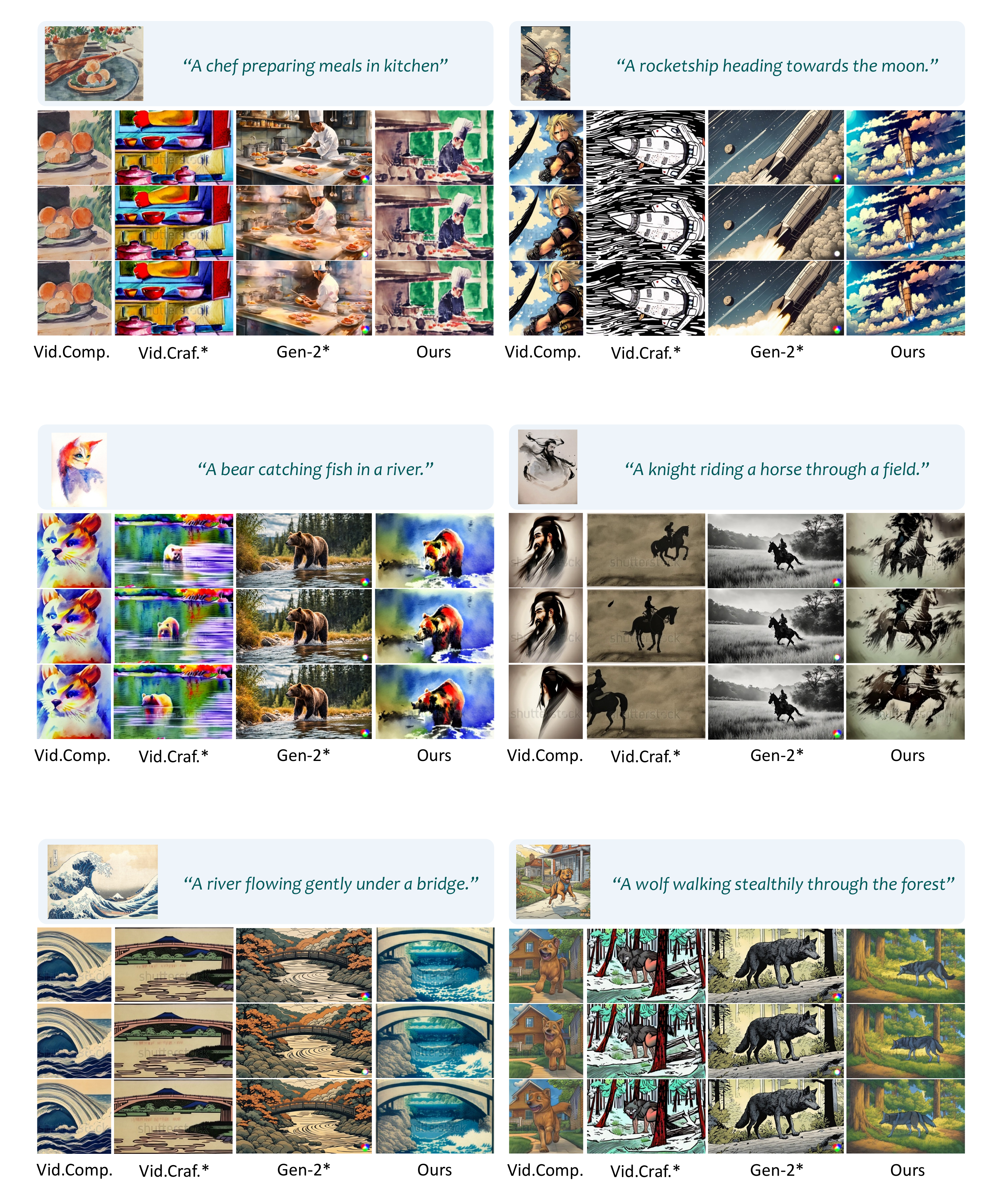}
    \caption{More Visual Comparison on Sinlge-Reference Stylized T2V Generation. Vid.Comp.: VideoComposer; Vid.Craf.: VideoCrafter.} 
    \label{fig:supp_more_result_comp_single}
\end{figure*}

\begin{figure*}[t]
    \centering
    \vspace{-1.5em}
    \includegraphics[width=0.85\linewidth]{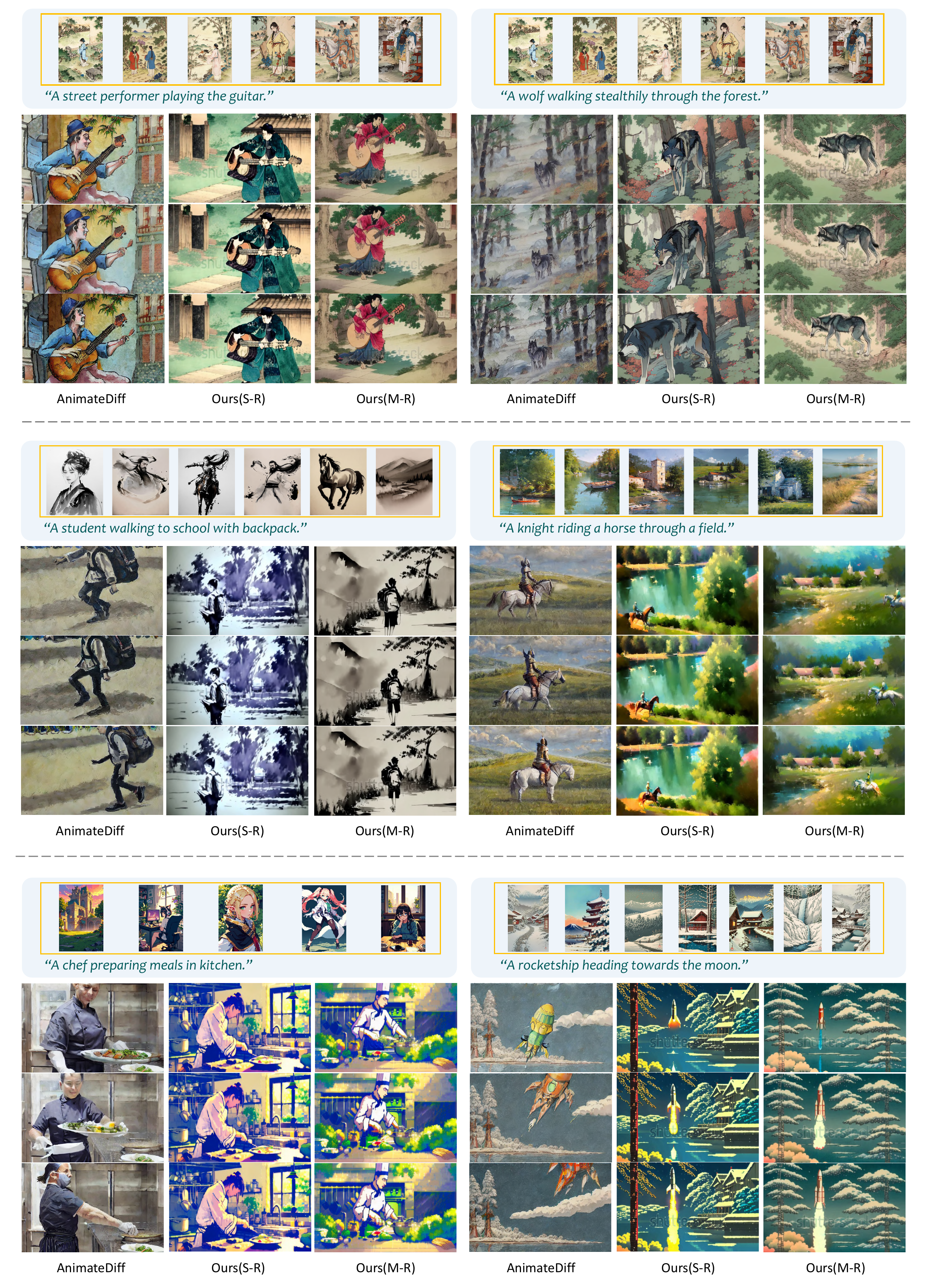}
    \vspace{-0.8em}
    \caption{More Visual Comparison on Multi-Reference Stylized T2V Generation} 
    \label{fig:supp_more_result_comp_multi}
\end{figure*}

\begin{figure*}[t]
    \centering
    \includegraphics[width=0.9\linewidth]{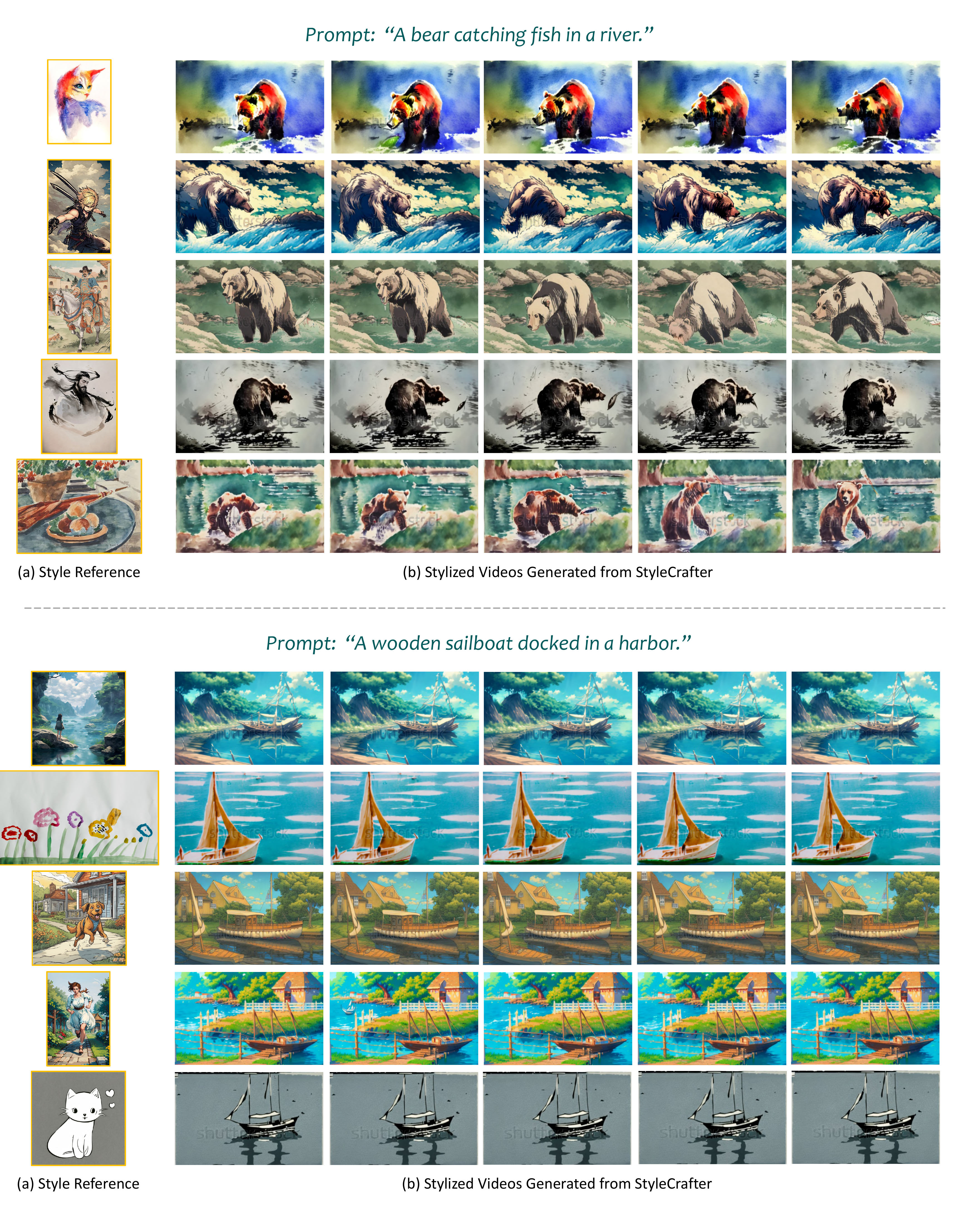}
    \vspace{-1.5em}
    \caption{More Results of StyleCrafter on Style-Guided Text-to-Video Generation} 
    \label{fig:supp_more_result_ours_1}
\end{figure*}

\begin{figure*}[t]
    \centering
    \includegraphics[width=0.9\linewidth]{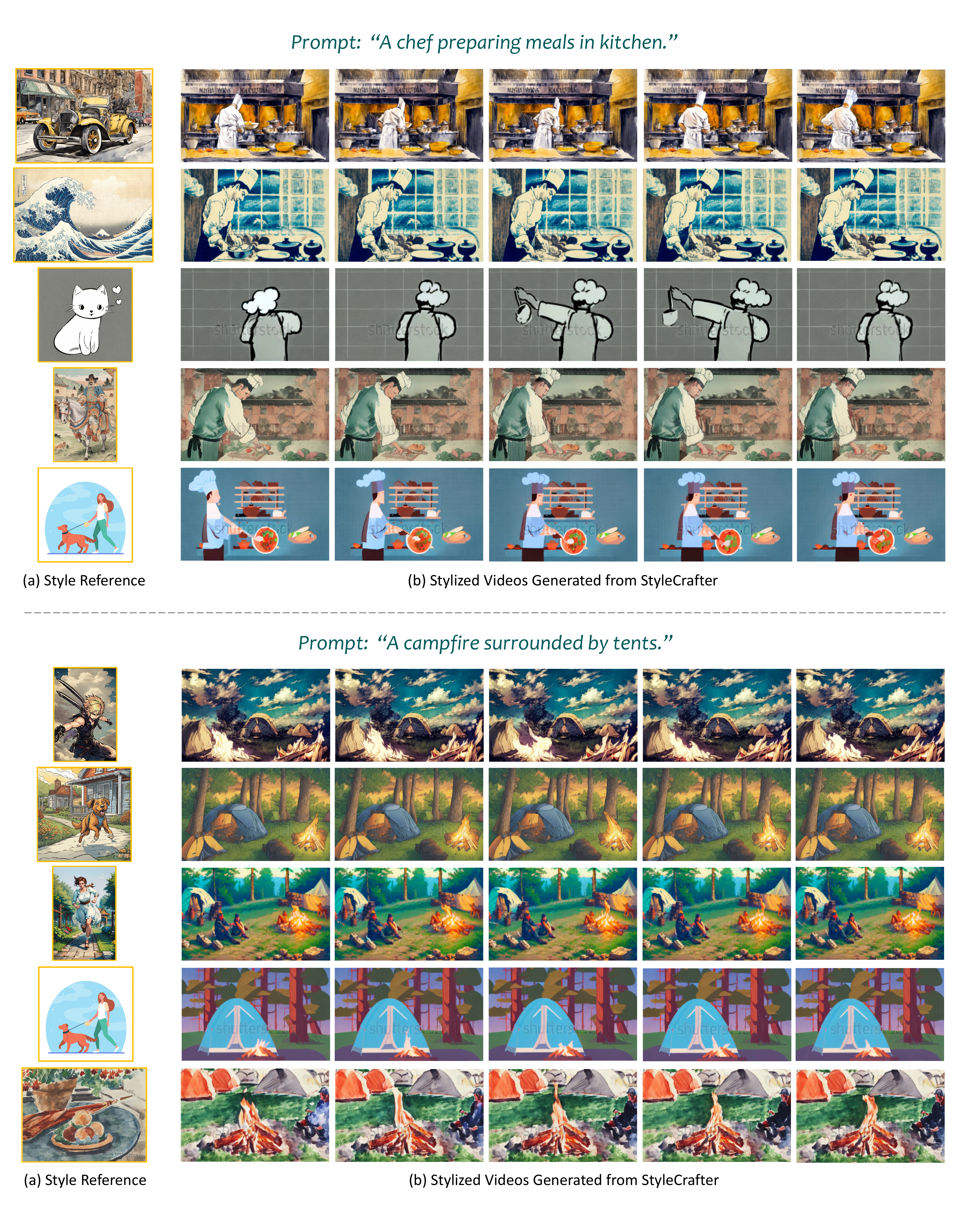}
    \vspace{-1em}
    \caption{More Results of StyleCrafter on Style-Guided Text-to-Video Generation} 
    \label{fig:supp_more_result_ours_2}
\end{figure*}

In this section, we provide more visual results and comparisons of our method. Specifically, we provide: (i) style-guided text-to-image generation results on StyleCrafter(SD2) and StyleCrafter(SDXL) in Figure~\ref{fig:supp_more_result_img_sd} and Figure~\ref{fig:supp_more_result_img_sdxl}; (ii) comparison of single-reference stylized video generation and multi-reference stylized video generation, as illustrated in Figure~\ref{fig:supp_more_result_comp_single} and Figure~\ref{fig:supp_more_result_comp_multi}, respectively. (iii) additional stylized video results in Figure~\ref{fig:supp_more_result_ours_1} and Figure~\ref{fig:supp_more_result_ours_2}.

%\clearpage

\section{Limitations}
\label{sec:supp_limitation}

\begin{figure}[!h]
    \centering
    \includegraphics[width=\linewidth]{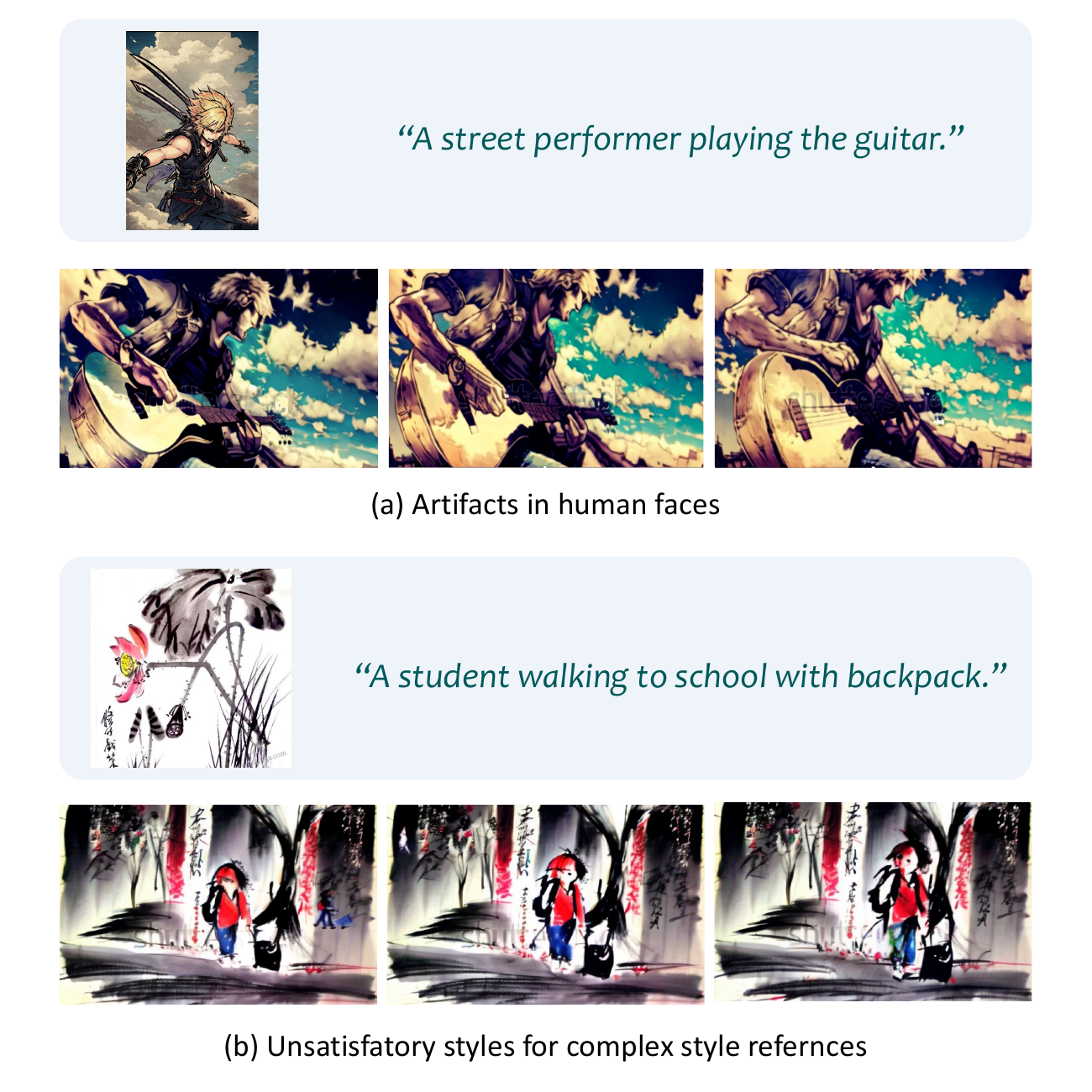}
    \caption{Failure cases of our methods. (a). Our method inherits limitations from the pretrained T2V model, tends to generate human faces with obvious artifacts. (b). Our method produces less satisifactory styles for complex style references such as Chinese Ink Painting.} 
    \label{fig:supp_failure}
\end{figure}

While our proposed method effectively handles most common styles, it does have certain limitations. Firstly, since StyleCrafter is developed based on existing T2I models and T2V models, such as SDXL and VideoCrafter, it unavoidably inherits part of the base model's shortcomings, such as fixed resolution and video length, less satisfactory temporal consistency. For example, our method fails to generate high-definition faces in certain cases, as shown in Figure~\ref{fig:supp_failure}. Despite the fact that our approach successfully enhances the stylistic generation capacity of T2I/T2V models, leveraging a powerful base model will always amplify its performance, as showcased in the superior style conformity of StyleCrafter(SDXL) over StyleCrafter(SD2).

Besides, artistic style is a very comprehensive perceptual feeling and visual styles are considerably more complex than what we explore in our paper. Our model may produce just passable results when confronted with reference images possessing highly stylized semantics. For example, as depicted in Figure~\ref{fig:supp_failure}, although our model successfully reproduces ink strokes, there are still discrepancies with reference images in the aesthetic level, such as the lack of "blank-leaving" in the generation results. Additionally, considering the absence of stylized video data, our stylized video generation results are somewhat less satisfactory than stylized image generation in visual style expression. A possible solution is to collect sufficient stylized video data for training, which we leave for further work.

%%%%%% for arxiv %%%%%% 

\end{sloppypar}
\end{document}